\documentclass[11pt]{article}

\usepackage[margin=1in]{geometry}

\usepackage{amsmath,amssymb,amsfonts}

\usepackage{algorithm}
\usepackage{algorithmic}

\usepackage{graphicx}
\usepackage{xcolor}

\usepackage{float}
\usepackage{placeins}

\usepackage{array}
\usepackage{booktabs}
\usepackage{multirow}
\usepackage{tabularx}
\usepackage{ragged2e}

\usepackage{url}
\usepackage[hidelinks]{hyperref}

\newcolumntype{L}[1]{>{\RaggedRight\arraybackslash}p{#1}}
\newcolumntype{Y}{>{\RaggedRight\arraybackslash}X}

\title{Self-Healing Agentic Orchestrators for Reliable Tool-Augmented Large Language Model Systems}

\author{
Rahul Suresh Babu\\
Independent Researcher\\
United States of America\\
\texttt{rahulsb@bu.edu}
\and
Adarsh Agrawal\\
Senior Member, IEEE\\
United States of America\\
\texttt{adagrawal@cs.stonybrook.edu}
}

\date{}

\begin{document}

\maketitle

\begin{abstract}
Tool-augmented large language model (LLM) agents rely on orchestration layers that coordinate planning, retrieval, tool invocation, validation, memory, and recovery. In these systems, failures arise not only from model errors, but also from orchestration-level issues such as tool timeouts, malformed arguments, stale context, contradictory evidence, retry loops, and unverified intermediate outputs. This paper presents a self-healing agentic orchestrator that treats reliability as a bounded runtime control problem. The orchestrator maps observable failure signals to inferred failure classes, selects targeted recovery actions under explicit budgets, verifies recovered trajectories, and records observability traces. We evaluate the approach on a 100-task controlled fault-injection benchmark against static workflow, retry-only, ReAct-style, and full-replanning baselines. Self-healing achieves 98.8\% task success, compared with 94.5\% for retry-only and 93.8\% for full replanning. A matched recovery-budget sweep shows that self-healing outperforms retry-only and full replanning at every tested budget, with the largest gap under a single recovery attempt: 94.0\% versus 85.3\% and 88.2\%, respectively. Under a controlled semantic silent-failure setting, verifier-guided self-healing reduces silent failures to 0.0\%, while non-verifying baselines return wrong-but-plausible outputs more often. A compact model-in-the-loop validation shows that the same recovery mechanism can operate when a live tool-calling model performs tool selection, argument generation, and answer synthesis over local fault-injected tools. These results provide controlled evidence that failure-aware, budgeted, and verification-guided orchestration improves reliability and diagnosability in tool-augmented LLM systems.
\end{abstract}

\noindent\textbf{Keywords:} Tool-augmented LLM agents, agentic orchestration, self-healing systems, LLM reliability, fault injection, runtime recovery, failure diagnosis, recovery policies, verification, observability.

\section{Introduction}
\label{sec:introduction}

Large language models (LLMs) are increasingly used as components of agentic systems that plan, retrieve information, invoke tools, maintain intermediate state, and generate final responses~\cite{lewis2020retrieval,yao2023react,schick2023toolformer,qin2024toolllm,wang2024surveyagents}. In such systems, reliability is not determined only by the base model. It also depends on the orchestration layer that coordinates tool calls, retrieved context, validation, recovery decisions, and termination. Failures can occur at the boundaries between the model and its execution environment: a tool may time out, a generated argument may violate a schema, retrieved evidence may be stale, two sources may disagree, a retry loop may make no progress, or an apparently successful answer may be unsupported by intermediate evidence~\cite{liu2024agentbench,winston2025taxonomy,gupta2026reliabilitybench}.

Common recovery mechanisms for LLM agents are often failure-agnostic. Static workflows terminate or follow fixed branches; retry mechanisms repeat failed calls; prompt-level self-correction asks the model to revise its behavior; and full replanning restarts execution from a broader planning step~\cite{yao2023react,shinn2023reflexion,liu2024agentbench,winston2025taxonomy}. These mechanisms are useful, but they do not explicitly connect observable failure signals to likely failure causes, targeted recovery actions, recovery budgets, and post-recovery verification. A timeout may justify retry with backoff, but stale retrieval context requires evidence refresh, malformed arguments require repair, contradictory evidence requires cross-checking, and wrong-but-plausible outputs require verification rather than another blind attempt.

This paper treats reliability in tool-augmented LLM systems as an orchestration-control problem. We propose a modular self-healing agentic orchestrator that implements a monitor--detect--diagnose--recover--verify loop. The orchestrator observes execution, detects failure signals, classifies likely failure causes, selects recovery actions conditioned on execution state and remaining recovery budget, verifies recovered trajectories, and records observability traces. The key design principle is that recovery should be targeted and bounded: the system should preserve useful execution state when possible, choose recovery actions appropriate to the inferred failure class, and avoid unbounded retry or replanning loops.

We formalize self-healing orchestration using execution states, failure events, failure classes, recovery actions, recovery budgets, and reliability objectives. We then evaluate the approach in two complementary settings. First, we use a controlled 100-task fault-injection benchmark with deterministic tools and matched fault schedules to isolate orchestration behavior. This benchmark compares self-healing against static workflow, retry-only, ReAct-style, and full-replanning baselines under runtime, tool-output, context, and semantic fault conditions. Second, we add a compact model-in-the-loop validation in which a live tool-calling model performs tool selection, argument generation, and answer synthesis while tools remain local, deterministic, and fault-injected. This second setting is not a full production-API evaluation, but it provides a bridge between deterministic orchestration experiments and live model behavior.

The controlled experiments show that self-healing improves reliability under the evaluated fault models. In the main controlled benchmark, self-healing reaches 98.8\% task success, compared with 94.5\% for retry-only and 93.8\% for full replanning. A recovery-budget sensitivity analysis shows that the improvement is not merely due to making more attempts: with a single recovery attempt, self-healing achieves 94.0\% success, compared with 85.3\% for retry-only and 88.2\% for full replanning, and it remains strongest at every matched budget. In the controlled semantic silent-failure setting, verifier-guided self-healing reduces silent failures to 0.0\%, while non-verifying baselines return wrong-but-plausible outputs more often. The model-in-the-loop validation further shows that the same recovery-aware orchestration mechanism can operate when a live model participates in tool use and answer synthesis.

The main contributions of this paper are as follows:
\begin{itemize}
    \item We formulate self-healing orchestration as a system-level reliability problem for tool-augmented LLM agents, defining execution states, failure signals, failure classes, recovery actions, recovery budgets, and reliability objectives.

    \item We propose a modular self-healing orchestrator that separates execution monitoring, failure detection, root-cause classification, recovery-policy selection, post-recovery verification, and observability.

    \item We introduce a targeted, budgeted recovery policy that maps inferred failure classes to recovery actions such as retry, argument repair, tool substitution, retrieval refresh, replanning, graceful degradation, and escalation.

    \item We evaluate the approach using a controlled fault-injection benchmark with static workflow, retry-only, ReAct-style, and full-replanning baselines, including failure-type analysis and ablations of classifier, verifier, targeted recovery, budget, and observability components.

    \item We add a controlled recovery-budget sensitivity experiment showing that targeted self-healing outperforms retry-only and full replanning under matched recovery budgets, especially when recovery opportunities are constrained.

    \item We provide compact model-in-the-loop validation and semantic silent-failure analysis, showing how verifier-guided recovery affects wrong-but-plausible outputs and how the orchestration loop behaves when a live tool-calling model interacts with local fault-injected tools.
\end{itemize}

The remainder of the paper formalizes the reliability problem, presents the self-healing orchestrator, describes the controlled and model-in-the-loop evaluations, reports empirical results, and discusses limitations and deployment implications.

\section{Background and Related Work}
\label{sec:background}

\subsection{Tool-Augmented LLM Agents}
\label{subsec:tool_augmented_agents}

Tool-augmented LLM agents extend standalone language models with retrieval, external APIs, code execution, memory, and interaction with external environments. This line of work builds on retrieval-augmented generation, which combines parametric models with non-parametric retrieval~\cite{lewis2020retrieval}; reasoning-and-acting methods such as ReAct, which interleave intermediate reasoning with environment actions~\cite{yao2023react}; and tool-use methods such as Toolformer and ToolLLM, which study when and how models invoke external tools and incorporate tool outputs~\cite{schick2023toolformer,qin2024toolllm}. Surveys of LLM-based agents further organize this space around agent construction, planning, memory, tool use, applications, and evaluation~\cite{wang2024surveyagents}.

These systems expand the capabilities of LLMs, but they also introduce new reliability dependencies. A tool-augmented agent must not only generate plausible text; it must select appropriate tools, construct valid arguments, handle external outputs, maintain context, and decide whether intermediate evidence is reliable enough to support subsequent actions. Failures therefore arise not only from model reasoning, but also from tool invocation, schema mismatch, stale or contradictory context, invalid execution state, weak validation, and orchestration decisions. This paper focuses on this orchestration layer: the control logic that determines whether an agent can detect, recover from, and explain failures during multi-step tool-augmented execution.

\subsection{Recovery Patterns in Agentic Systems}
\label{subsec:recovery_patterns}

Existing agentic systems use several recovery mechanisms, but they usually address only part of the runtime reliability problem. Static workflows and graph-based controllers can encode deterministic fallback paths, but they often lack explicit failure diagnosis. Retry mechanisms, fallbacks, and circuit-breaker patterns are effective for transient infrastructure failures, but they do not distinguish timeouts from stale context, malformed arguments, incorrect tool selection, or semantic inconsistency. ReAct-style loops can adapt through observation and action, while reflection-based systems such as Reflexion use feedback to improve future decisions~\cite{yao2023react,shinn2023reflexion}. Planner-executor and full-replanning approaches can recover from some planning errors, but they may discard useful partial state and introduce unnecessary latency or cost when local repair would be sufficient.

This creates a recovery-mismatch problem: the selected recovery action may not address the underlying failure cause. Retrying can help a transient timeout, but it does not repair stale retrieval context; full replanning may address an invalid plan, but it can be excessive when a local argument repair, tool substitution, or retrieval refresh would suffice. A reliability-oriented orchestrator should therefore condition recovery on the observed failure signal, inferred failure class, current execution state, and remaining recovery budget.

The proposed self-healing orchestrator is related to autonomic-computing control loops such as MAPE-K, which separate monitoring, analysis, planning, execution, and knowledge for self-managing systems~\cite{kephart2003vision,ibm2006autonomic}. However, tool-augmented LLM agents introduce agent-specific failure signals and recovery decisions: malformed tool arguments, verifier rejection, stale retrieval context, contradictory evidence, repeated reasoning/tool-use loops, and unsupported final answers. Our focus is therefore not a general autonomic architecture, but an agentic orchestration policy that explicitly maps observable failure signals to inferred failure classes, targeted recovery actions, recovery budgets, post-recovery verification, and traceable execution records.

\begin{table}[t]
\centering
\caption{Comparison of recovery mechanisms for tool-augmented LLM systems. ``Budget'' denotes explicit bounds on recovery attempts, latency, cost, or recovery depth.}
\label{tab:recovery_mechanism_comparison}
\begin{tabular}{lcccccc}
\hline
Approach & Detect & Diagnose & Targeted & Budget & Verify & Trace \\
\hline
Static workflow & -- & -- & Limited & -- & Optional & Limited \\
Retry / fallback / circuit breaker & Yes & -- & -- & Limited & -- & Limited \\
ReAct-style loop & Implicit & -- & Implicit & -- & -- & Partial \\
Full replanning & Yes & -- & Coarse & Limited & Optional & Partial \\
MAPE-K-style control & Yes & Yes & Depends & Depends & Depends & Yes \\
Self-healing orchestration & Yes & Yes & Yes & Yes & Yes & Yes \\
\hline
\end{tabular}
\end{table}

Table~\ref{tab:recovery_mechanism_comparison} summarizes this distinction. The goal of self-healing orchestration is not to replace retry, replanning, fallback, circuit breaking, or verification mechanisms. Instead, it treats them as recovery actions within a broader control policy that selects the appropriate action based on the observed failure signal, inferred root cause, current execution state, and remaining recovery budget.

\subsection{Reliability Evaluation and Fault Injection}
\label{subsec:reliability_evaluation}

Agent evaluation has increasingly moved beyond single-turn language generation toward interactive, tool-using, and multi-step task settings. AgentBench evaluates LLMs as agents across multiple interactive environments and highlights failures related to long-horizon reasoning, decision-making, and instruction following~\cite{liu2024agentbench}. Tool-augmented failure analyses emphasize that errors may arise from tool selection, argument construction, tool execution, output interpretation, and downstream reasoning~\cite{winston2025taxonomy}. Reliability-oriented benchmarks further argue that single-run task success is insufficient for production-oriented agents because it does not capture consistency, robustness to perturbations, or fault tolerance under infrastructure failures~\cite{gupta2026reliabilitybench}.

This paper follows that reliability-oriented view but focuses specifically on orchestration-level recovery. We use controlled fault injection to isolate how different orchestration policies behave under matched tool, context, output, and semantic failure conditions. Controlled evaluation provides internal validity: each method encounters the same tasks, fault schedules, and success criteria. We complement this with compact model-in-the-loop validation, where a live tool-calling model performs tool selection, argument generation, and answer synthesis while tools remain local and fault-injected. Together, these settings separate two questions: whether targeted recovery improves reliability under controlled faults, and whether the same orchestration mechanism remains usable when a live model participates in tool-augmented execution.

\section{Problem Formulation}
\label{sec:problem}

This section formulates self-healing orchestration as a bounded runtime reliability problem for tool-augmented LLM agents. In this paper, \emph{self-healing} does not mean unrestricted autonomous repair. It refers to recovery within predefined failure classes, recovery actions, verification checks, and execution budgets. The formulation abstracts agent execution as a sequence of states, actions, observations, failure signals, recovery decisions, and verification outcomes. It does not assume that all failures are directly observable or that inferred failure classes are always correct. Instead, it provides a control-oriented view for selecting and evaluating recovery actions under bounded cost, latency, and recovery-depth constraints.

\subsection{Execution Model}
\label{subsec:task_execution_model}

Let a user task be represented as $T$. The agent executes a trajectory
\begin{equation}
\tau = \{(S_1,a_1,o_1), (S_2,a_2,o_2), \ldots, (S_n,a_n,o_n)\},
\label{eq:trajectory}
\end{equation}
where $S_t$ is the orchestrator state at step $t$, $a_t$ is the selected action, and $o_t$ is the observation returned after executing the action. An action may correspond to model reasoning, retrieval, memory access, tool invocation, validation, replanning, recovery, escalation, or final response generation. An observation may include a tool response, retrieved evidence, validation result, execution error, latency measurement, verifier output, or response candidate.

The state $S_t$ contains the information needed to continue execution or recover from failure: the task, current plan, action history, intermediate observations, tool outputs, retrieved context, verifier feedback, retry history, recovery history, budget state, and execution metadata. This representation treats the orchestrator as a runtime control layer over tool-augmented execution rather than only as a prompt template, tool selector, or static workflow.

\subsection{Failure Signals, Events, and Classes}
\label{subsec:failure_events}

A failure event $F_t$ occurs when executing action $a_t$ violates an expected execution condition. Some failures are explicit, such as tool timeouts, execution exceptions, authentication errors, rate limits, invalid schemas, or malformed tool outputs. Others are implicit, such as contradictory evidence, low verifier confidence, repeated actions, lack of progress, unsupported conclusions, or a final response that is not grounded in intermediate observations.

We distinguish among failure signals, failure events, and failure classes. Let $x_t$ denote an observable signal associated with a possible failure at step $t$, such as an exception, timeout, schema violation, verifier rejection, contradiction, or repeated-action pattern. Let $F_t$ denote the corresponding failure event when the observed behavior violates an expected condition. Let $C_t$ denote the inferred failure class, such as tool invocation failure, schema or argument failure, tool-output failure, context failure, planning failure, verification failure, or control-loop failure. The class $C_t$ may not be directly observable and must be inferred from the signal $x_t$ and execution state $S_t$.

A detected failure occurs when the orchestrator observes a failure signal and initiates recovery. A silent failure occurs when execution returns an incorrect, unsupported, stale, or inconsistent result without detection. This distinction is central to the evaluation: detectable runtime faults test whether the orchestrator can recover from explicit execution failures, while semantic silent-failure settings test whether verification can detect wrong-but-plausible outputs that do not raise ordinary runtime errors.

\subsection{Recovery Policy and Budget}
\label{subsec:recovery_actions}

Given a detected failure event $F_t$, the orchestrator selects a recovery action $R_t$ from a set of allowable actions:
\begin{equation}
\begin{aligned}
\mathcal{R} = \{&
\mathrm{retry}, \mathrm{repair}, \mathrm{substitute}, \mathrm{replan},\\
&\mathrm{refresh}, \mathrm{escalate}, \mathrm{degrade}, \mathrm{terminate}\},\\
R_t &\in \mathcal{R}.
\end{aligned}
\label{eq:recovery_action_set}
\end{equation}

The recovery action is selected by a policy
\begin{equation}
R_t = \pi(S_t, x_t, C_t, B_t),
\label{eq:recovery_policy}
\end{equation}
where $S_t$ is the current orchestrator state, $x_t$ is the observed failure signal, $C_t$ is the inferred failure class, and $B_t$ is the remaining recovery budget. The recovery budget is represented as
\begin{equation}
B_t = (b_t^{r}, b_t^{l}, b_t^{k}, b_t^{d}),
\label{eq:recovery_budget}
\end{equation}
where $b_t^{r}$ is the remaining recovery-attempt budget, $b_t^{l}$ is the remaining latency budget, $b_t^{k}$ is the remaining cost budget, and $b_t^{d}$ is the remaining recovery-depth budget. These constraints prevent unbounded retry storms, excessive replanning, uncontrolled tool use, and recovery actions whose expected cost exceeds their expected reliability benefit.

After applying a recovery action, the orchestrator obtains a verification outcome $v_t$ and updates the execution state:
\begin{equation}
S_{t+1} = \Phi(S_t, a_t, o_t, R_t, v_t),
\label{eq:state_update}
\end{equation}
where $\Phi$ denotes the state-update function that incorporates the prior state, executed action, observation, recovery decision, and verification result. The same abstraction supports local recovery, such as argument repair or retry, and broader recovery, such as retrieval refresh, replanning, graceful degradation, or escalation.

\subsection{Reliability Objectives}
\label{subsec:reliability_objectives}

The objective is to improve task reliability under failure conditions while bounding the additional cost and latency introduced by recovery. For a benchmark containing $N$ task executions, task success rate is defined as
\begin{equation}
\mathrm{TSR} = \frac{1}{N}\sum_{i=1}^{N}\mathbb{I}[\text{task } i \text{ is completed correctly}],
\label{eq:tsr}
\end{equation}
where correctness is determined by the task-specific success criteria defined in the experimental setup.

For the subset of detected failures, recovery success rate is defined as
\begin{equation}
\mathrm{RSR} =
\frac{\#\text{successfully recovered detected failures}}
{\#\text{detected failures}}.
\label{eq:rsr}
\end{equation}

Silent failure rate is defined as
\begin{equation}
\mathrm{SFR} =
\frac{\#\text{incorrect outputs returned without detection}}
{\#\text{total completed outputs}}.
\label{eq:sfr}
\end{equation}
We report silent failure rate over completed outputs because silent failures are specifically failures that appear externally successful.

In addition to reliability metrics, recovery policies must respect latency and cost constraints. We use a general cost model
\begin{equation}
K = \alpha M + \beta U + \gamma V + \delta R,
\label{eq:cost_proxy_general}
\end{equation}
where $M$ is the number of model calls, $U$ is the number of tool calls, $V$ is the number of verifier calls, $R$ is the number of recovery actions, and $\alpha,\beta,\gamma,\delta$ are normalized unit-cost weights. The controlled experiments instantiate this general model using a simple call-count cost proxy, described in Section~\ref{sec:experimental_setup}. This separates the formal reliability objective from provider-specific monetary pricing or production latency, which depend on deployment configuration.

\section{Self-Healing Orchestrator}
\label{sec:architecture}

\subsection{Architecture Overview}
\label{subsec:architecture_overview}

\begin{figure}[t]
    \centering
    \includegraphics[width=\linewidth]{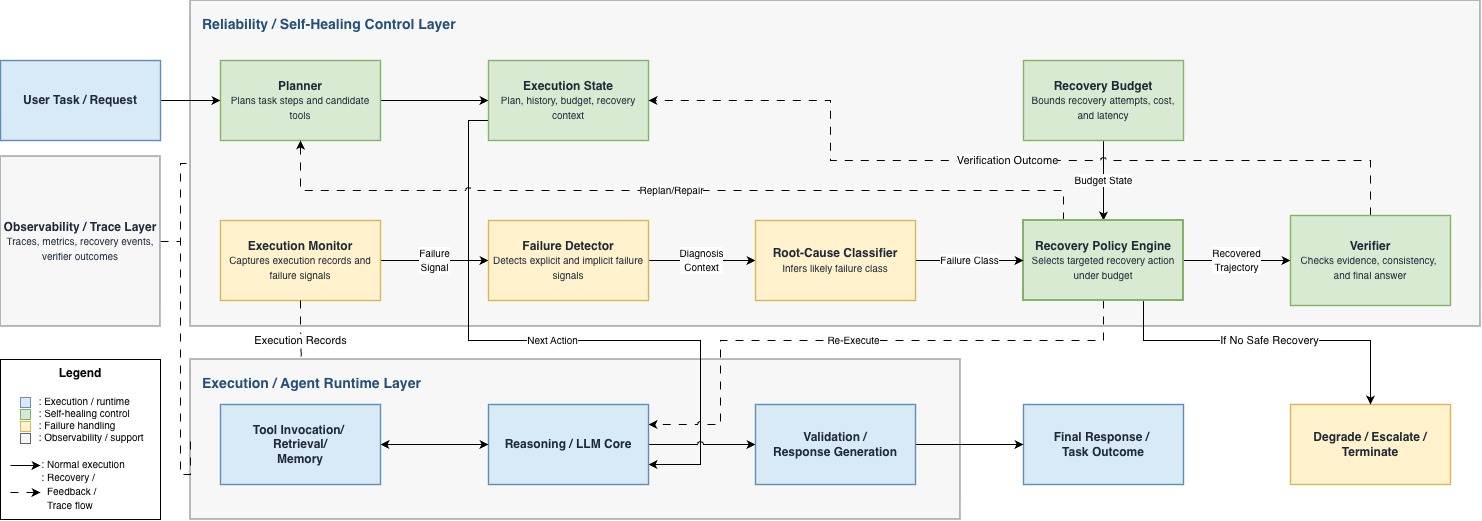}
    \caption{Self-healing orchestrator architecture. A reliability control plane surrounds the agent execution data plane, mapping failure signals to targeted recovery actions, verifying recovered trajectories, and emitting traces for diagnosis and evaluation.}
    \label{fig:self_healing_architecture}
\end{figure}

Figure~\ref{fig:self_healing_architecture} illustrates the proposed self-healing orchestrator. The architecture separates the agent execution data plane from a reliability control plane. The data plane performs ordinary tool-augmented execution: model reasoning, retrieval, memory access, tool invocation, validation, and response generation. The control plane monitors this execution, detects failure signals, diagnoses likely failure classes, selects recovery actions under an explicit budget, verifies recovered trajectories, and records observability traces.

This separation is important because failures in tool-augmented LLM systems often occur at the boundary between model reasoning and external execution. Even when the underlying model is capable, the system can fail because a tool is unavailable, an argument is malformed, retrieved context is stale, evidence is contradictory, a verifier rejects the response, or a recovery loop fails to terminate. The self-healing orchestrator addresses these failures as runtime control problems: it preserves execution state, identifies the likely failure class, applies a bounded recovery action, and verifies whether execution can safely continue.

The design is modular. Different deployments may replace the planner, detector, classifier, verifier, or recovery policy with domain-specific implementations while preserving the same control-loop structure. The controlled benchmark instantiates these components with deterministic tools and interpretable recovery rules to isolate orchestration behavior, while the model-in-the-loop validation evaluates the same control structure when a live tool-calling model drives tool selection, argument generation, and answer synthesis over local fault-injected tools.

\subsection{Control-Plane Components}
\label{subsec:control_plane_components}

The control plane consists of seven logical components. The planner proposes executable actions from the current task and state. The execution monitor records the action, observation, tool metadata, latency, validation results, and budget state. The failure detector converts execution evidence into candidate failure signals. The root-cause classifier maps a detected signal and execution state to an inferred failure class. The recovery policy engine implements the policy $R_t = \pi(S_t, x_t, C_t, B_t)$ from Section~\ref{sec:problem}. The verifier evaluates whether a recovered trajectory or final response is valid and task-aligned. The observability layer records the resulting execution and recovery trace.

Table~\ref{tab:architecture_interfaces} summarizes the primary interfaces among these components. The table is intentionally aligned with the notation in Section~\ref{sec:problem}: state $S_t$, action $a_t$, observation $o_t$, failure signal $x_t$, failure class $C_t$, recovery action $R_t$, recovery budget $B_t$, and verification outcome $v_t$.

\begin{table}[t]
\centering
\caption{Primary component interfaces in the self-healing orchestrator.}
\label{tab:architecture_interfaces}
\small
\begin{tabularx}{\textwidth}{L{0.18\textwidth} L{0.27\textwidth} L{0.27\textwidth} Y}
\toprule
\textbf{Component} & \textbf{Primary Inputs} & \textbf{Primary Outputs} & \textbf{Role} \\
\midrule
Planner & Task $T$, state $S_t$, constraints, verifier feedback & Plan $P_t$, candidate action $a_t$, replanning constraints & Proposes executable next steps \\
\midrule
Execution monitor & Action $a_t$, observation $o_t$, tool metadata, latency, errors & Execution record, updated metadata, candidate signal $x_t$ & Captures structured execution evidence \\
\midrule
Failure detector & Execution record, validation results, verifier feedback, retry history & Failure signal $x_t$ or no-failure decision & Detects deviation from expected behavior \\
\midrule
Root-cause classifier & State $S_t$, signal $x_t$, execution history, validation results & Inferred failure class $C_t$ & Estimates likely failure cause \\
\midrule
Recovery policy engine & State $S_t$, signal $x_t$, class $C_t$, budget $B_t$ & Recovery action $R_t$, budget update & Selects a bounded recovery action \\
\midrule
Verifier & Task $T$, recovered state, tool outputs, retrieved evidence, response candidate & Verification outcome $v_t$, accept/reject decision & Checks validity of recovered execution \\
\midrule
Observability layer & Execution events, signals, classes, actions, budgets, verifier outcomes & Traces, metrics, logs, final task status & Makes recovery diagnosable and evaluable \\
\bottomrule
\end{tabularx}
\end{table}

The components are separated to avoid conflating detection, diagnosis, and recovery. The detector answers whether execution has deviated from expected behavior. The classifier estimates why the deviation occurred. The recovery policy then chooses what to do under the remaining budget. This separation allows the orchestrator to handle heterogeneous failures more precisely than a uniform retry or full-replanning strategy.

\subsection{Verification and Observability}
\label{subsec:verification_observability}

Verification is treated as a first-class control-plane component rather than as a final formatting check. The verifier may check schema validity, output completeness, consistency with retrieved evidence, and support for the final response. If verification fails, the rejection is converted into a recoverable signal and passed back into the same control loop. This is especially important for semantic silent failures: a tool output or final answer may be syntactically valid and appear successful while remaining stale, unsupported, contradictory, or incorrect. By treating verifier rejection as a failure signal, the orchestrator can recover from wrong-but-plausible outputs that ordinary runtime monitoring may not detect.

Observability records the decisions made by the control plane. Each trace can include the task state, tool calls, retrieved evidence, failure signals, inferred failure classes, selected recovery actions, verifier outcomes, budget consumption, and final status. These traces do not merely support post-hoc debugging; they also enable failure analysis, benchmark construction, regression testing, recovery-policy tuning, and comparison across orchestration strategies. In the experiments, trace records are used both to evaluate recovery behavior quantitatively and to inspect representative recovery cases qualitatively.

Overall, the architecture implements the monitor--detect--diagnose--recover--verify loop defined in Section~\ref{sec:problem}. The next section instantiates this architecture with a concrete failure-to-recovery mapping and budgeted recovery policy.

\section{Recovery Policy Design}
\label{sec:policy}

This section instantiates the abstract recovery policy $R_t = \pi(S_t,x_t,C_t,B_t)$ from Section~\ref{sec:problem}. The policy is intentionally interpretable: it maps inferred failure classes to recovery actions, applies those actions under explicit budgets, verifies recovered state, and records trace events. The objective is to choose the lowest-cost feasible recovery action that is likely to restore valid execution, rather than applying a uniform retry or full-replanning strategy to every failure.

\subsection{Failure-to-Recovery Mapping}
\label{subsec:failure_recovery_mapping}

Table~\ref{tab:failure_recovery_mapping} summarizes the policy template used in the experiments. The mapping is not intended to be exhaustive; it defines an interpretable set of recovery actions for the failure classes studied in this paper. Domain-specific deployments may add failure classes, change thresholds, or restrict recovery actions for safety, authorization, compliance, or irreversible side-effect constraints.

\begin{table}[t]
\centering
\caption{Failure-to-recovery mapping for self-healing orchestration.}
\label{tab:failure_recovery_mapping}
\small
\begin{tabularx}{\textwidth}{L{0.18\textwidth} L{0.25\textwidth} L{0.31\textwidth} Y}
\toprule
\textbf{Failure class} & \textbf{Representative signals} & \textbf{Primary recovery actions} & \textbf{Escalation or termination condition} \\
\midrule
Tool timeout or unavailable tool & Timeout, unavailable endpoint, execution exception, rate limit & Retry with backoff; substitute equivalent tool; replan around unavailable dependency & Repeated timeout, rate-limit exhaustion, or no substitute available \\
\midrule
Schema or argument failure & Invalid JSON, missing field, type mismatch, rejected arguments & Repair arguments; regenerate structured call; apply schema-constrained prompt & Repeated invalid arguments or unsupported schema \\
\midrule
Malformed or incomplete output & Empty output, partial response, parse failure, missing fields & Validate output; request corrected output; use alternate tool & Output remains invalid after repair or alternate call \\
\midrule
Incorrect tool selection & Verifier rejects tool choice, irrelevant output, unsupported operation & Replan current step; select alternative tool; constrain tool-selection candidates & Repeated tool-selection failure \\
\midrule
Stale or insufficient context & Low evidence score, outdated retrieved content, missing context & Refresh retrieval; query alternate source; update memory context & Evidence remains insufficient or outdated \\
\midrule
Contradictory evidence & Conflicting tool outputs or retrieved documents & Cross-check with independent source; invoke verifier; request clarification & Conflict cannot be resolved within budget \\
\midrule
Verification or semantic failure & Verifier rejection, unsupported claim, low evidence score, wrong-but-plausible output & Retrieve supporting evidence; regenerate with evidence constraints; add verification step & Verification remains below threshold \\
\midrule
Control-loop failure & Retry storm, repeated actions, no progress, excessive replanning & Terminate loop; replan from stable checkpoint; degrade; escalate & Maximum recovery budget reached \\
\bottomrule
\end{tabularx}
\end{table}

The policy treats retry, repair, substitution, retrieval refresh, replanning, verification, degradation, and escalation as recovery actions rather than as separate orchestration paradigms. This is the central difference from recovery mechanisms that apply the same action to every failure. The selected action depends on the observed signal, inferred failure class, current execution state, and remaining recovery budget.

\subsection{Recovery Budget}
\label{subsec:recovery_budget}

For the implemented policy, we instantiate the abstract budget from Section~\ref{sec:problem} using counters for retry, replanning, tool substitution, model escalation, latency, and cost:
\begin{equation}
B_t = (b_t^{r}, b_t^{p}, b_t^{s}, b_t^{m}, b_t^{L}, b_t^{K}),
\label{eq:policy_budget}
\end{equation}
where $b_t^{r}$ is the remaining retry budget, $b_t^{p}$ is the remaining replanning budget, $b_t^{s}$ is the remaining tool-substitution budget, $b_t^{m}$ is the remaining model-escalation budget, $b_t^{L}$ is the remaining latency budget, and $b_t^{K}$ is the remaining cost budget.

The budget serves three roles: it prevents unbounded retry or replanning loops, makes reliability-cost trade-offs explicit, and defines when the system should degrade, escalate, or terminate. After each recovery action, the budget is updated as
\begin{equation}
B_{t+1} = B_t - \Delta B(R_t),
\label{eq:policy_budget_update}
\end{equation}
where $\Delta B(R_t)$ is the budget consumed by the selected recovery action. Failed recovery attempts still consume budget because their latency and cost have already been incurred. Section~\ref{sec:results} evaluates whether targeted recovery remains beneficial when retry-only, full replanning, and self-healing are given matched recovery budgets.

\subsection{Policy Algorithm}
\label{subsec:policy_algorithm}

Algorithm~\ref{alg:self_healing_policy} shows the high-level policy. The algorithm separates ordinary execution from recovery: normal actions update state directly, while detected failures or verifier rejections are routed through classification, budgeted recovery, verification, and trace logging.

\begin{algorithm}[t]
\caption{Self-Healing Orchestration Policy}
\label{alg:self_healing_policy}
\small
\begin{algorithmic}[1]
\STATE Initialize task $T$, plan $P$, execution state $S_1$, recovery budget $B_1$, and trace log $\mathcal{G}$
\STATE Set $t \leftarrow 1$
\WHILE{task is not complete and $B_t$ is not exhausted}
    \STATE Select next action $a_t$ from plan $P$ and state $S_t$
    \STATE Execute action $a_t$ and observe output $o_t$
    \STATE Record $(a_t,o_t)$ in trace log $\mathcal{G}$
    \STATE Update state: $S_t \leftarrow \mathrm{Update}(S_t,a_t,o_t)$
    \STATE Detect failure signal $x_t$
    \IF{no failure signal is detected}
        \IF{task completion condition is satisfied}
            \STATE Verify candidate final state and obtain $v_t$
            \IF{$v_t$ is accepted}
                \STATE \textbf{return} final response and trace log $\mathcal{G}$
            \ELSE
                \STATE Set $x_t \leftarrow$ verifier rejection
            \ENDIF
        \ELSE
            \STATE $t \leftarrow t + 1$
            \STATE Continue execution
        \ENDIF
    \ENDIF
    \IF{failure signal $x_t$ is present}
        \STATE Classify likely failure class $C_t$
        \STATE Select recovery action $R_t = \pi(S_t,x_t,C_t,B_t)$
        \IF{$R_t$ is terminate or no feasible action exists under $B_t$}
            \STATE \textbf{break}
        \ENDIF
        \STATE Apply recovery action $R_t$ and observe recovery outcome
        \STATE Verify recovered state and obtain verification outcome $v_t$
        \STATE Update budget: $B_{t+1} = B_t - \Delta B(R_t)$
        \STATE Update state $S_{t+1}$ using recovery outcome and $v_t$
        \STATE Record $(x_t,C_t,R_t,v_t,B_{t+1})$ in trace log $\mathcal{G}$
        \STATE $t \leftarrow t + 1$
    \ENDIF
\ENDWHILE
\IF{task completed and verified}
    \STATE Return final response and trace log $\mathcal{G}$
\ELSIF{budget is exhausted and partial result is safe to return}
    \STATE Return degraded response with uncertainty or missing-dependency notice
\ELSE
    \STATE Escalate or terminate execution
\ENDIF
\end{algorithmic}
\end{algorithm}

The experimental implementation uses fixed recovery mappings and deterministic budget updates so that differences between methods can be attributed to orchestration behavior rather than nondeterministic policy variation. The same structure could also support learned or model-assisted recovery policies, but this paper uses an interpretable policy for reproducibility and analysis.

\subsection{Escalation and Policy Scope}
\label{subsec:policy_scope}

When multiple recovery actions are feasible, the policy prioritizes local, low-cost repairs before global recovery. Argument repair, retry, retrieval refresh, and tool substitution are preferred when the failure is narrow and the remaining state is still trustworthy. Replanning, graceful degradation, model or human escalation, and termination are reserved for failures that cannot be repaired locally or whose recovery budget has been exhausted. This priority ordering avoids both extremes: blindly retrying cheap but ineffective actions and prematurely invoking expensive full replanning.

\section{Experimental Setup}
\label{sec:experimental_setup}

We evaluate whether self-healing orchestration improves reliability under tool, context, recovery, and semantic failure conditions. The evaluation has two complementary tracks. First, a controlled fault-injection benchmark isolates orchestration behavior under deterministic tools, fixed task definitions, matched fault schedules, and paired random seeds. This setting provides internal validity: each method encounters the same tasks, faults, success criteria, and execution limits. Second, a compact model-in-the-loop validation tests whether the same recovery mechanism remains usable when a live tool-calling model performs tool selection, argument generation, and answer synthesis over local fault-injected tools. The model-in-the-loop setting is not a production-API evaluation; tools remain local and deterministic, and faults remain controlled.

\subsection{Research Questions}
\label{subsec:research_questions}

The experiments are designed to answer the following questions:
\begin{itemize}
    \item \textbf{RQ1:} Does self-healing orchestration improve task success under controlled runtime and tool-output faults?
    \item \textbf{RQ2:} Which injected failure classes benefit most from targeted recovery?
    \item \textbf{RQ3:} Which self-healing components contribute most to reliability?
    \item \textbf{RQ4:} Does targeted recovery outperform retry-only and full replanning under matched recovery budgets?
    \item \textbf{RQ5:} Does verifier-guided recovery reduce semantic silent failures?
    \item \textbf{RQ6:} Does the same orchestration mechanism operate when a live tool-calling model drives tool interaction over local fault-injected tools?
\end{itemize}

\subsection{Controlled Benchmark}
\label{subsec:controlled_benchmark}

The controlled benchmark contains 100 synthetic tool-augmented tasks designed to exercise planning, tool selection, retrieval, computation, evidence checking, and recovery behavior. The tasks are divided evenly across five categories: retrieval and evidence synthesis, multi-step API workflows, calculation and verification, planning and tool selection, and contradiction resolution. Each task is represented as a structured record containing a task identifier, category, user request, required tools, expected answer, success criteria, eligible faults, and execution plan. Before fault-injection experiments, all tasks are validated in the no-fault setting to ensure that failures observed during evaluation are caused by injected faults or orchestration behavior rather than malformed task definitions.

The controlled environment uses deterministic tools with known ground-truth behavior. The tool set includes document search, policy lookup, customer lookup, order lookup, inventory lookup, calculator, shipping estimator, evidence checker, and tool-registry functions. Each tool exposes a structured schema and returns either a valid response or an injected failure response according to the fault schedule. Simulated tools are used deliberately: they make fault timing, recovery opportunities, and success criteria reproducible across methods.

\subsection{Fault Injection}
\label{subsec:fault_injection}

The controlled benchmark injects detectable runtime and tool-output faults at eligible tool steps. We consider seven fault classes: timeout, unavailable tool, malformed output, schema drift, partial response, stale context, and contradictory evidence. Faults are sampled using an intensity parameter
\begin{equation}
\lambda \in \{0.0,0.1,0.2,0.3,0.4,0.5\}.
\end{equation}
The range $\lambda \in \{0.0,0.1,0.2,0.3\}$ is treated as the primary regime, while $\lambda \in \{0.4,0.5\}$ is used as a high-stress regime. Because $\lambda$ is applied per eligible tool step rather than per task, multi-step tasks have a higher probability of encountering at least one injected fault.

Fault schedules are deterministic and task-aware. The seed for each potential fault depends on the task identifier, method, trial seed, recovery attempt, step index, and fault intensity. This makes runs reproducible while preventing all tasks from sharing identical fault patterns. The same seed set is shared across compared methods, enabling paired comparison under matched fault conditions.

Semantic faults are evaluated separately. In the semantic silent-failure setting, a correct tool output may be replaced by a wrong-but-plausible output that is syntactically valid but violates the task-specific success criterion. This setting tests whether an orchestrator can detect unsupported or incorrect outputs that do not raise ordinary runtime errors.

\subsection{Compared Methods}
\label{subsec:baselines}

The controlled experiments compare five orchestration strategies:
\begin{itemize}
    \item \textbf{Static workflow:} executes a predefined tool sequence without adaptive recovery.
    \item \textbf{Retry-only:} retries failed execution using a fixed retry limit, without failure classification or targeted recovery.
    \item \textbf{ReAct-style:} simulates repeated reasoning-and-acting steps inspired by ReAct~\cite{yao2023react}, but without explicit recovery budgets, failure-class diagnosis, or verifier-guided recovery.
    \item \textbf{Full replanning:} discards the failed attempt and replans before re-execution.
    \item \textbf{Self-healing:} detects failure signals, classifies likely failure causes, selects targeted recovery actions under budget, verifies recovered trajectories, and records traces.
\end{itemize}

All controlled methods use the same tasks, tools, fault-injection protocol, success criteria, timeout thresholds, and execution limits. Method-specific differences are restricted to orchestration behavior: whether the method retries, replans, classifies failure causes, invokes verification, applies targeted recovery, or records recovery traces. The model-in-the-loop experiments focus on the recovery-enabled subset of methods used in the live-model setting: retry-only, full replanning, and self-healing.

\subsection{Model-in-the-Loop Validation}
\label{subsec:model_in_loop_setup}

The controlled benchmark abstracts model behavior to isolate orchestration effects. To test whether the same mechanism remains usable with live model behavior, we add a compact model-in-the-loop validation. In this setting, a live tool-calling model performs tool selection, structured argument generation, and final answer synthesis. Tools remain local and deterministic, and faults are injected using the same reliability-oriented principles as in the controlled benchmark. This design isolates live model/tool interaction without introducing uncontrolled production API behavior, external service availability, or naturally occurring production incidents.

The model-in-the-loop evaluation contains three parts: a main validation comparing retry-only, full replanning, and self-healing over 90 executions; a verifier ablation comparing verifier-on and verifier-off self-healing over 60 executions; and a supplemental budget-sensitivity sweep over 180 executions. These experiments are intentionally smaller than the controlled benchmark. Their purpose is not to replace the controlled fault-injection study, but to provide an external-validity bridge showing that the orchestration mechanism can run when a live model participates in tool use and answer synthesis.

\subsection{Evaluation Metrics}
\label{subsec:evaluation_metrics}

We report reliability, recovery, efficiency, and diagnosability metrics. Task success rate, recovery success rate, and silent failure rate follow the definitions in Section~\ref{subsec:reliability_objectives}. We also report average recovery steps, tool calls, model calls, verifier calls, simulated latency, escalation rate, and trace completeness. For controlled experiments, cost is reported using a simple call-count proxy:
\begin{equation}
\mathrm{CostProxy} = N_{\mathrm{tool}} + N_{\mathrm{model}} + N_{\mathrm{verify}},
\end{equation}
where $N_{\mathrm{tool}}$, $N_{\mathrm{model}}$, and $N_{\mathrm{verify}}$ denote tool, model/planning, and verifier calls. This proxy compares relative orchestration overhead rather than monetary cost. Model-in-the-loop artifacts additionally record call and token metadata where available, but we do not treat these as production dollar-cost estimates because monetary cost depends on provider pricing, model choice, prompt length, verifier implementation, and deployment configuration.

\subsection{Experiment Suite}
\label{subsec:experiment_summary}

Table~\ref{tab:experiment_summary} summarizes the full experiment suite. The controlled experiments are designed to isolate mechanism-level reliability under matched faults. The model-in-the-loop experiments provide a compact validation of the same orchestration loop with live model participation.

\begin{table}[t]
\centering
\caption{Summary of experimental design.}
\label{tab:experiment_summary}
\footnotesize
\begin{tabularx}{\textwidth}{L{0.25\textwidth} L{0.42\textwidth} L{0.13\textwidth} Y}
\toprule
\textbf{Experiment} & \textbf{Design} & \textbf{Executions} & \textbf{Primary metrics} \\
\midrule
Controlled main reliability 
& 100 tasks $\times$ 5 methods $\times$ 6 fault intensities $\times$ 3 seeds 
& 9000 
& Task success, recovery, cost, latency \\
\midrule
Failure-class analysis 
& Faulted subset of controlled main experiment, grouped by injected fault class and method 
& Subset 
& Success and recovery by fault class \\
\midrule
Controlled ablation 
& 100 tasks $\times$ 6 self-healing variants $\times$ 5 nonzero fault intensities $\times$ 3 seeds 
& 9000 
& Success, recovery, cost, traces \\
\midrule
Controlled budget sensitivity 
& 100 tasks $\times$ 3 methods $\times$ 4 budgets $\times$ 2 fault intensities $\times$ 3 seeds 
& 7200 
& Success versus budget and cost \\
\midrule
Semantic silent-failure 
& 100 tasks $\times$ 6 semantic variants $\times$ 4 semantic fault intensities $\times$ 3 seeds 
& 7200 
& Silent failure, success, verifier calls \\
\midrule
Trace diagnosability 
& Representative recovered executions from self-healing runs 
& 3 traces 
& Signal, class, action, verifier outcome \\
\midrule
Model-in-loop main 
& 15 tasks $\times$ 3 methods $\times$ 2 seeds 
& 90 
& Success, detection, recovery, traces \\
\midrule
Model-in-loop verifier ablation 
& 15 tasks $\times$ 2 verifier variants $\times$ 2 seeds 
& 60 
& Success, detection, recovery \\
\midrule
Model-in-loop budget sensitivity 
& 10 tasks $\times$ 3 methods $\times$ 3 budgets $\times$ 2 seeds 
& 180 
& Success and recovery by budget \\
\bottomrule
\end{tabularx}
\end{table}

\subsection{Controlled Experiments}
\label{subsec:controlled_experiments}

The controlled main reliability experiment evaluates RQ1 by comparing static workflow, retry-only, ReAct-style, full replanning, and self-healing across six runtime fault intensities and three deterministic seeds. Results are aggregated overall and broken down by fault intensity and task category.

The failure-class analysis evaluates RQ2 by restricting analysis to executions in which at least one fault was injected. Results are grouped by injected fault type and method to determine where targeted recovery provides the largest benefit.

The ablation study evaluates RQ3 by comparing full self-healing with variants that remove or relax individual components: without verifier, without root-cause classifier, relaxed recovery budget, without targeted recovery, and without observability traces. This determines whether reliability gains come from the complete self-healing loop or from a single component.

The controlled budget-sensitivity experiment evaluates RQ4 by matching recovery budgets across retry-only, full replanning, and self-healing. Budgets are swept over $\{1,2,3,5\}$ at fault intensities $\lambda \in \{0.2,0.4\}$. This experiment tests whether targeted recovery remains beneficial when all recovery-enabled methods are given the same number of recovery opportunities.

The semantic silent-failure experiment evaluates RQ5 using wrong-but-plausible outputs. Semantic fault intensity is swept over
\begin{equation}
\lambda_s \in \{0.0,0.1,0.2,0.3\}.
\end{equation}
This experiment compares baselines, verifier-guided self-healing, and no-verifier self-healing to test whether verification reduces undetected semantic errors.

The trace diagnosability analysis extracts three representative self-healing traces: a runtime/tool failure, a context/evidence failure, and a semantic silent-failure case. These traces qualitatively evaluate whether the orchestrator records the failure signal, inferred class, recovery action, verifier outcome, budget use, and final status.

\subsection{Execution Protocol}
\label{subsec:experimental_protocol}

All controlled methods are evaluated under matched task, tool, fault, and seed schedules. Tool schemas, task definitions, success criteria, timeout thresholds, maximum execution steps, semantic fault schedules, and recovery limits are fixed unless a baseline explicitly requires different behavior. Aggregate results are reported across all tasks, with breakdowns by method, fault intensity, task category, injected failure class, ablation variant, budget level, and semantic fault intensity.

The model-in-the-loop experiments use the same artifact discipline: fixed task sets, fixed local tools, controlled fault schedules, repeated seeds, structured traces, and health-check summaries. The controlled benchmark should therefore be interpreted as the primary mechanism-isolation study, while the model-in-the-loop experiments provide a compact bridge to live model behavior. Neither setting claims to measure production incident rates, external API reliability, or deployment-specific monetary cost.

\section{Results}
\label{sec:results}

This section reports the controlled benchmark results followed by compact model-in-the-loop validation. The controlled experiments isolate orchestration behavior under deterministic tools, matched fault schedules, and fixed success criteria. The model-in-the-loop experiments test whether the same recovery mechanism operates when a live tool-calling model performs tool selection, argument generation, and answer synthesis over local fault-injected tools. Unless otherwise stated, percentages are reported as mean values over the repeated deterministic trials described in Section~\ref{sec:experimental_setup}.

\subsection{Controlled Reliability}
\label{subsec:controlled_reliability}

Table~\ref{tab:overall_results} reports aggregate performance across 100 tasks, six runtime fault intensities, and three seeds. Self-healing achieves the highest overall task success rate, reaching 98.8\%, compared with 94.5\% for retry-only, 94.1\% for the ReAct-style baseline, 93.8\% for full replanning, and 70.1\% for static workflow. This is a 4.3 percentage-point improvement over the strongest non-self-healing baseline, retry-only, and a 5.0 percentage-point improvement over full replanning.

\begin{table}[t]
\centering
\caption{Overall controlled reliability comparison across orchestration strategies.}
\label{tab:overall_results}
\small
\setlength{\tabcolsep}{6pt}
\begin{tabular}{lcccccc}
\toprule
\textbf{Method} 
& \textbf{Success} 
& \textbf{Recovery} 
& \textbf{Silent} 
& \textbf{Tool} 
& \textbf{Verifier} 
& \textbf{Cost} \\
& \textbf{(\%)} 
& \textbf{(\%)} 
& \textbf{Failure (\%)} 
& \textbf{Calls} 
& \textbf{Calls} 
& \textbf{Proxy} \\
\midrule
Static workflow & 70.1 & 0.0  & 0.0 & 1.61 & 0.00 & 1.61 \\
Retry-only      & 94.5 & 24.0 & 0.0 & 2.24 & 0.00 & 2.24 \\
ReAct-style     & 94.1 & 22.6 & 0.0 & 2.23 & 0.00 & 3.63 \\
Full replanning & 93.8 & 21.0 & 0.0 & 2.23 & 0.00 & 3.63 \\
Self-healing    & 98.8 & 27.6 & 0.0 & 2.25 & 1.00 & 3.25 \\
\bottomrule
\end{tabular}
\end{table}

Static workflow has the lowest cost proxy because it does not attempt adaptive recovery, but this also makes it the least reliable under injected faults. Retry-only is a strong low-cost baseline, showing that simple recovery helps with some transient failures. However, self-healing reaches higher reliability while using a lower cost proxy than ReAct-style and full replanning. The zero silent-failure rate in Table~\ref{tab:overall_results} should be interpreted only within the detectable runtime-fault setting; semantic wrong-but-plausible outputs are evaluated separately in Section~\ref{subsec:semantic_silent_failure_results}.

\begin{figure}[t]
    \centering
    \includegraphics[width=\linewidth]{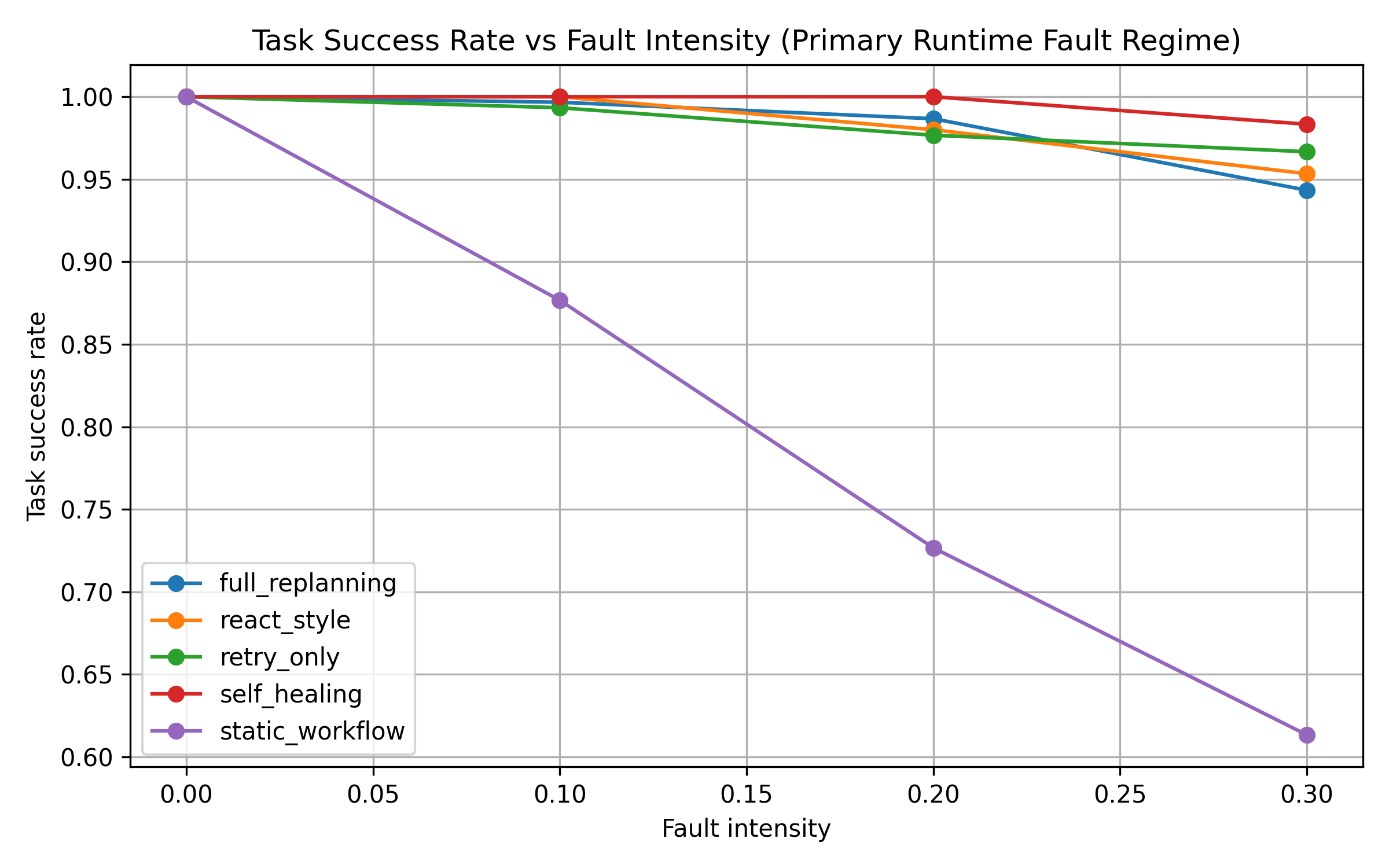}
    \caption{Task success rate under increasing runtime fault intensity. Self-healing degrades least as fault intensity increases in the controlled benchmark.}
    \label{fig:success_vs_fault_intensity}
\end{figure}

Figure~\ref{fig:success_vs_fault_intensity} shows success as runtime fault intensity increases. All methods perform well in the no-fault setting, validating the task definitions and deterministic tool environment. As fault intensity increases, static workflow degrades sharply, while recovery-enabled methods degrade more gradually. Self-healing maintains the highest success rate across both the primary and high-stress regimes.

Table~\ref{tab:primary_stress_results} separates the primary regime, $\lambda \in \{0.0,0.1,0.2,0.3\}$, from the high-stress regime, $\lambda \in \{0.4,0.5\}$. In the primary regime, self-healing achieves 99.6\% success. In the high-stress regime, self-healing maintains 97.3\% success, while retry-only, ReAct-style, and full replanning fall to 86.7\%, 85.7\%, and 85.2\%, respectively. This suggests that targeted recovery becomes more valuable as tool instability increases.

\begin{table}[t]
\centering
\caption{Primary and high-stress controlled runtime fault results.}
\label{tab:primary_stress_results}
\small
\begin{tabular}{lcccc}
\toprule
\textbf{Method} & \textbf{Primary Success (\%)} & \textbf{Primary Cost} & \textbf{Stress Success (\%)} & \textbf{Stress Cost} \\
\midrule
Static workflow & 80.4 & 1.65 & 49.3 & 1.52 \\
Retry-only & 98.4 & 2.04 & 86.7 & 2.65 \\
ReAct-style & 98.3 & 3.26 & 85.7 & 4.38 \\
Full replanning & 98.2 & 3.28 & 85.2 & 4.32 \\
Self-healing & 99.6 & 3.03 & 97.3 & 3.68 \\
\bottomrule
\end{tabular}
\end{table}

\begin{figure}[t]
    \centering
    \includegraphics[width=\linewidth]{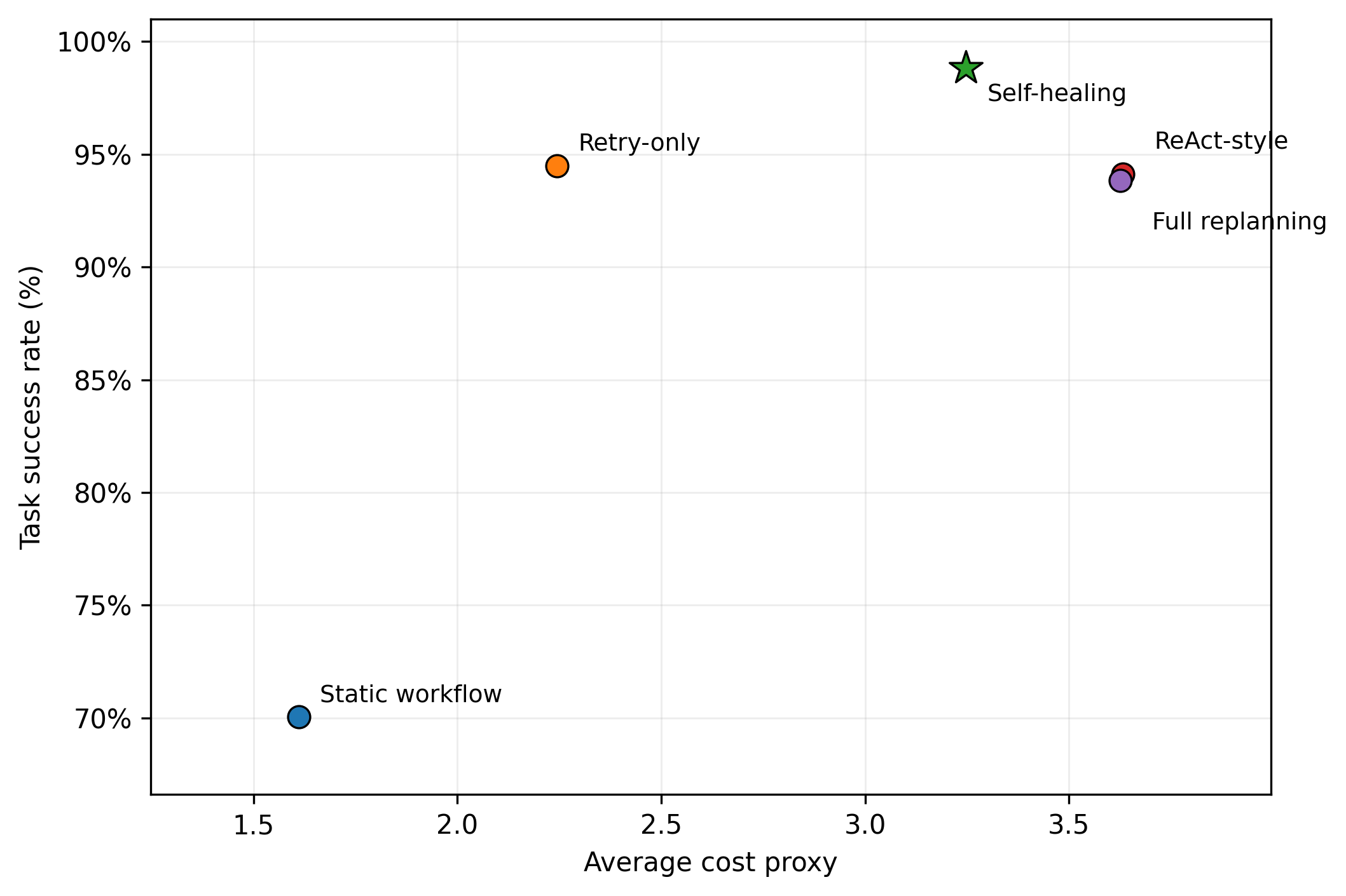}
    \caption{Cost--reliability trade-off across orchestration strategies in the controlled benchmark. Self-healing achieves the highest task success while using a lower cost proxy than ReAct-style and full replanning.}
    \label{fig:cost_reliability_tradeoff}
\end{figure}

The cost--reliability trade-off is shown in Figure~\ref{fig:cost_reliability_tradeoff}. Self-healing improves reliability over full replanning while reducing the cost proxy from 3.63 to 3.25. These controlled costs are relative call-count proxies, not provider-specific dollar costs. In a deployment, verifier cost and latency would depend on the model, prompt length, evidence context, tool latency, and verifier implementation.

\subsection{Failure-Type Analysis}
\label{subsec:failure_type_analysis}

Table~\ref{tab:failure_type_results} reports task success over the harder subset of executions in which at least one runtime or tool-output fault was actually injected. Static workflow has 0.0\% success in this subset because it has no recovery mechanism once an injected fault interrupts execution. Self-healing achieves the highest success for every evaluated fault class, supporting the design choice of mapping failure classes to targeted recovery actions.

\begin{table}[t]
\centering
\caption{Task success rate by injected failure type, computed only over executions with at least one injected fault.}
\label{tab:failure_type_results}
\small
\begin{tabular}{lccccc}
\toprule
\textbf{Failure Type} & \textbf{Static} & \textbf{Retry-Only} & \textbf{ReAct} & \textbf{Full Replan} & \textbf{Self-Healing} \\
\midrule
Contradictory evidence & 0.0 & 85.7 & 79.5 & 68.6 & 100.0 \\
Malformed output & 0.0 & 68.8 & 81.1 & 71.1 & 98.2 \\
Partial response & 0.0 & 59.0 & 61.1 & 61.1 & 89.7 \\
Schema drift & 0.0 & 62.7 & 49.6 & 62.8 & 89.6 \\
Stale context & 0.0 & 73.1 & 79.4 & 64.6 & 100.0 \\
Timeout & 0.0 & 62.9 & 55.3 & 52.9 & 79.3 \\
Unavailable tool & 0.0 & 64.4 & 50.9 & 44.0 & 75.3 \\
\bottomrule
\end{tabular}
\end{table}

\begin{figure}[t]
    \centering
    \includegraphics[width=\linewidth]{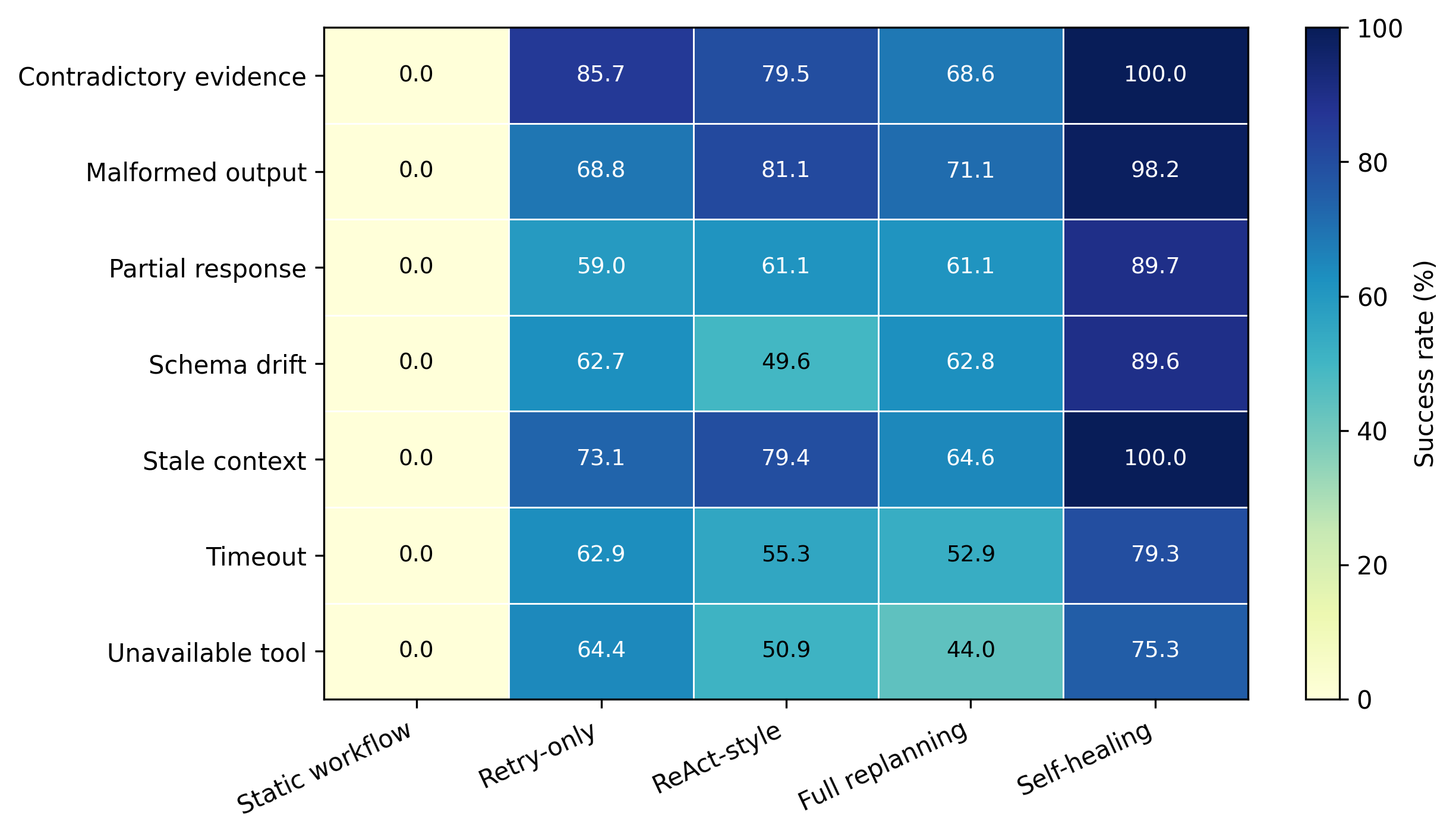}
    \caption{Task success rate by injected failure type and orchestration method. Self-healing is strongest across all evaluated controlled runtime fault classes.}
    \label{fig:failure_type_heatmap}
\end{figure}

The largest gains occur for stale context and contradictory evidence, where self-healing reaches 100.0\% success within the evaluated controlled fault templates. These values should not be interpreted as universal guarantees for arbitrary context or contradiction failures; they show that, under the benchmark's controlled templates, targeted retrieval refresh, evidence cross-checking, and verification are more effective than uniform retry or full replanning. Figure~\ref{fig:failure_type_heatmap} visualizes the same pattern.

\subsection{Ablation Study}
\label{subsec:ablation_study}

Table~\ref{tab:ablation_results} reports the controlled ablation study over nonzero runtime fault intensities. Removing the root-cause classifier reduces task success from 99.3\% to 95.7\%. Removing targeted recovery produces the same reduction, indicating that the reliability gain is not simply due to repeated attempts; the mapping from failure class to recovery action is important. Removing observability traces does not reduce task success, but it eliminates trace events, confirming that observability primarily supports diagnosability rather than direct task completion in this benchmark.

\begin{table}[t]
\centering
\caption{Ablation study results for the self-healing orchestrator under nonzero controlled runtime fault intensities.}
\label{tab:ablation_results}
\footnotesize
\setlength{\tabcolsep}{4.5pt}
\begin{tabularx}{\textwidth}{L{0.30\textwidth} c c c c c}
\toprule
\textbf{Variant} 
& \textbf{Success} 
& \textbf{Recovery} 
& \textbf{Recovery} 
& \textbf{Cost} 
& \textbf{Trace} \\
& \textbf{(\%)} 
& \textbf{(\%)} 
& \textbf{Steps} 
& \textbf{Proxy} 
& \textbf{Events} \\
\midrule
Full self-healing & 99.3 & 32.3 & 0.43 & 3.30 & 2.18 \\
No verifier & 99.3 & 32.3 & 0.43 & 2.30 & 2.18 \\
No classifier & 95.7 & 28.7 & 0.56 & 3.45 & 2.45 \\
Relaxed recovery budget & 100.0 & 33.0 & 0.44 & 3.31 & 2.21 \\
No targeted recovery & 95.7 & 28.7 & 0.56 & 3.45 & 2.45 \\
No observability traces & 99.3 & 32.3 & 0.43 & 3.30 & 0.00 \\
\bottomrule
\end{tabularx}
\end{table}

The no-verifier variant matches full self-healing in this detectable runtime-fault setting while reducing cost, because these faults usually expose explicit error signals. Verification becomes important in the semantic setting evaluated in Section~\ref{subsec:semantic_silent_failure_results}. The relaxed-budget variant reaches 100.0\% success, showing that additional recovery opportunities can increase completion. However, this result also raises a natural question: is targeted recovery better, or does simply trying more suffice? The next subsection evaluates this directly by matching recovery budgets across retry-only, full replanning, and self-healing.

\begin{figure}[t]
    \centering
    \includegraphics[width=\linewidth]{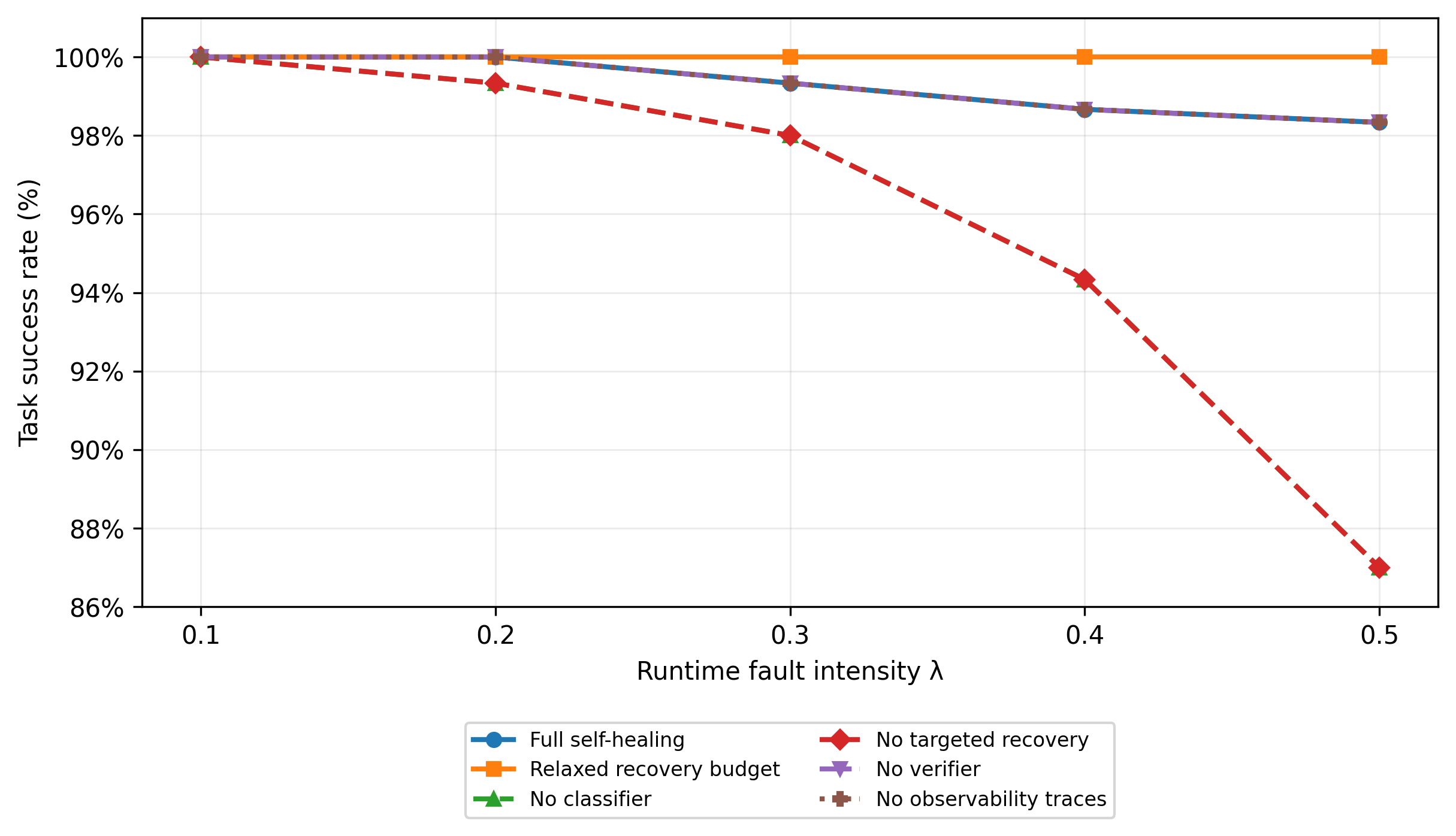}
    \caption{Ablation results under nonzero runtime fault intensities. Removing root-cause classification or targeted recovery reduces task success, indicating that recovery gains depend on failure-aware action selection.}
    \label{fig:ablation_success}
\end{figure}

\subsection{Recovery-Budget Sensitivity}
\label{subsec:budget_sensitivity_results}

Table~\ref{tab:budget_sensitivity_results} reports the controlled recovery-budget sensitivity experiment. Budgets are matched across retry-only, full replanning, and self-healing. More recovery budget improves all methods, but self-healing achieves the highest success at every matched budget. The largest gap appears when recovery opportunities are most constrained: with a single recovery attempt, self-healing reaches 94.0\% success, compared with 85.3\% for retry-only and 88.2\% for full replanning.

\begin{table}[t]
\centering
\caption{Controlled recovery-budget sensitivity. Values are task success rates (\%) under matched recovery budgets.}
\label{tab:budget_sensitivity_results}
\small
\begin{tabular}{lccc}
\toprule
\textbf{Recovery Budget} & \textbf{Retry-Only} & \textbf{Full Replanning} & \textbf{Self-Healing} \\
\midrule
1 & 85.3 & 88.2 & 94.0 \\
2 & 93.8 & 94.3 & 97.8 \\
3 & 96.7 & 97.8 & 99.2 \\
5 & 99.2 & 99.3 & 100.0 \\
\bottomrule
\end{tabular}
\end{table}

\begin{figure}[t]
    \centering
    \includegraphics[width=\linewidth]{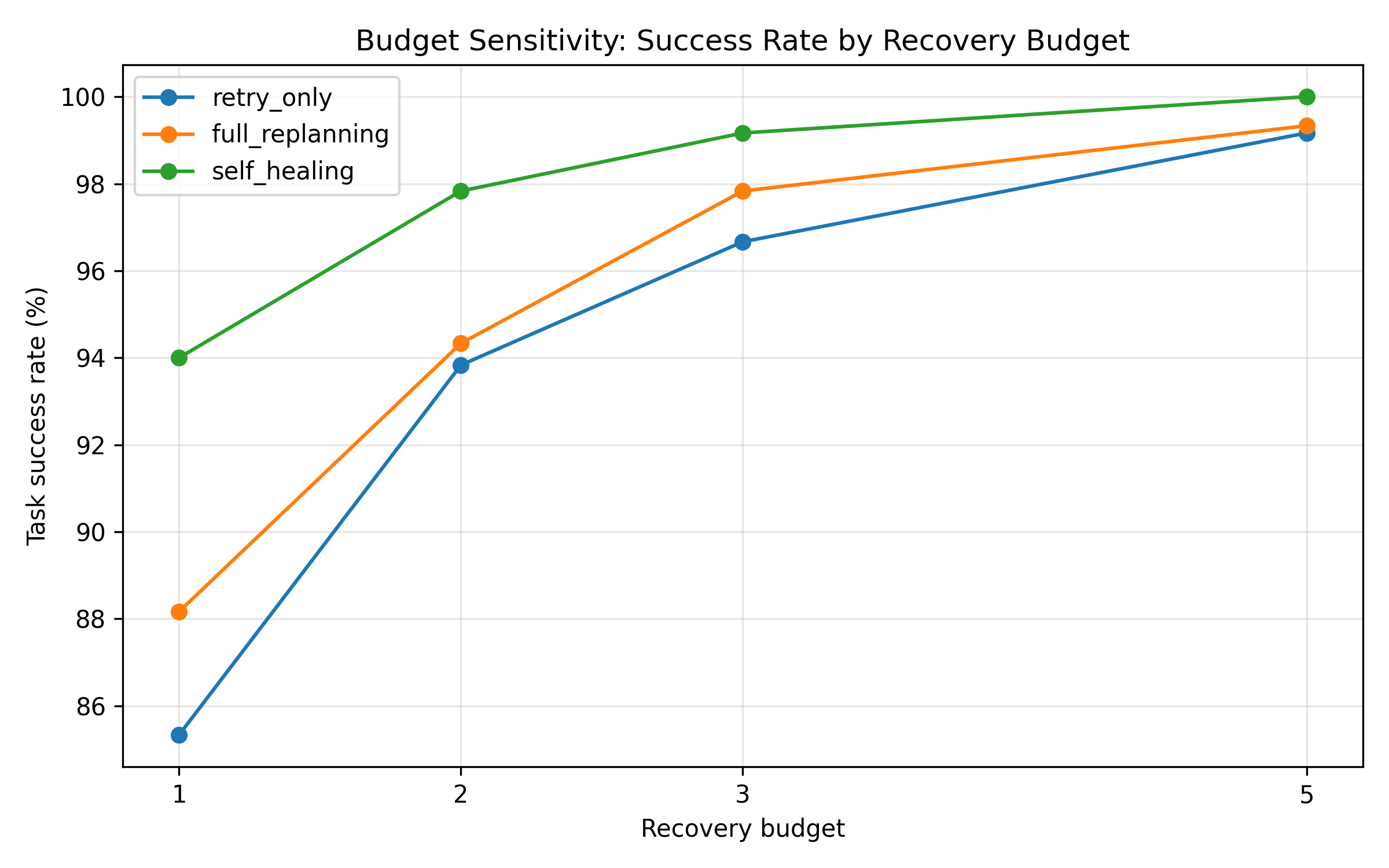}
    \caption{Controlled recovery-budget sensitivity. Self-healing reaches higher success at every matched recovery budget, with the largest advantage when recovery opportunities are constrained.}
    \label{fig:budget_success}
\end{figure}

This result directly addresses the ``just try more'' alternative. Additional attempts help, but targeted recovery uses limited recovery opportunities more effectively. The effect is strongest under higher fault intensity: at $\lambda=0.4$ and budget 1, self-healing reaches 90.3\% success, compared with 79.0\% for retry-only and 81.7\% for full replanning. Figure~\ref{fig:budget_success} shows the same trend across budgets.

The budget sweep also clarifies the reliability-cost trade-off. At budget 1, self-healing has a cost proxy of 3.13, compared with 2.15 for retry-only and 3.45 for full replanning. Thus, self-healing is more expensive than retry-only because it uses verification and targeted recovery logic, but it is less expensive than full replanning while achieving higher reliability. This supports the claim that targeted recovery improves reliability under bounded execution, rather than relying on unbounded retries or repeated global replanning.

\subsection{Semantic Silent-Failure Results}
\label{subsec:semantic_silent_failure_results}

The previous experiments focus primarily on detectable runtime and tool-output faults. Table~\ref{tab:semantic_silent_failure} reports the semantic silent-failure experiment, where tools return wrong-but-plausible outputs without explicit error signals. These outputs are syntactically valid and can appear successful to an orchestrator that only monitors runtime status.

\begin{table}[t]
\centering
\caption{Semantic silent-failure experiment under the controlled semantic fault model.}
\label{tab:semantic_silent_failure}
\footnotesize
\setlength{\tabcolsep}{4.5pt}
\begin{tabularx}{\textwidth}{L{0.34\textwidth} c c c c c}
\toprule
\textbf{Method} 
& \textbf{Success} 
& \textbf{Silent} 
& \textbf{Recovery} 
& \textbf{Verifier} 
& \textbf{Cost} \\
& \textbf{(\%)} 
& \textbf{Failure (\%)} 
& \textbf{(\%)} 
& \textbf{Calls} 
& \textbf{Proxy} \\
\midrule
Static workflow & 84.8 & 15.2 & 0.0 & 0.00 & 1.74 \\
Retry-only & 85.4 & 14.6 & 0.0 & 0.00 & 1.74 \\
ReAct-style & 86.8 & 13.2 & 0.0 & 0.00 & 2.74 \\
Full replanning & 84.5 & 15.5 & 0.0 & 0.00 & 2.74 \\
Self-healing without verifier & 82.4 & 17.6 & 0.0 & 0.00 & 1.74 \\
Self-healing with verifier & 100.0 & 0.0 & 13.9 & 1.14 & 3.12 \\
\bottomrule
\end{tabularx}
\end{table}

Verifier-guided self-healing achieves 100.0\% task success and 0.0\% silent failure under the evaluated controlled semantic fault model. In contrast, non-verifying methods return silent failures between 13.2\% and 17.6\% of the time. This result shows that runtime failure detection alone is insufficient for wrong-but-plausible outputs: semantic validation must be part of the recovery loop. The result should be interpreted within the controlled semantic fault templates and does not imply that verification eliminates all possible silent failures in open-world deployments.

\begin{figure}[t]
    \centering
    \includegraphics[width=\linewidth]{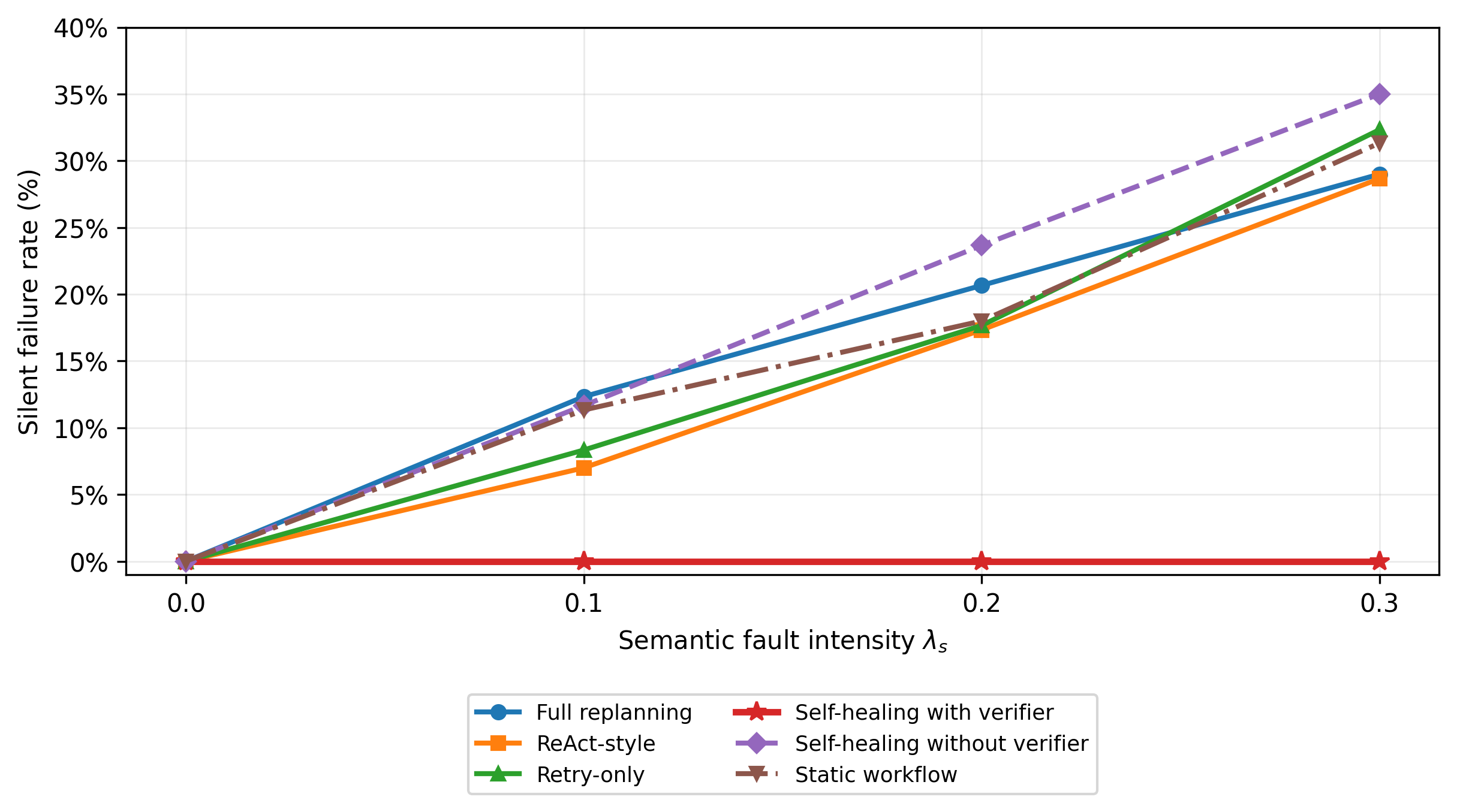}
    \caption{Silent failure rate under semantic wrong-output faults. Verifier-guided self-healing reduces silent failure to 0.0\% across the evaluated controlled semantic fault intensities.}
    \label{fig:semantic_silent_failure}
\end{figure}

\subsection{Model-in-the-Loop Validation}
\label{subsec:model_in_loop_results}

The controlled experiments isolate orchestration behavior under deterministic tools and fault schedules. To test whether the same mechanism remains usable with live model behavior, we evaluate a compact model-in-the-loop setting. In this setting, a live tool-calling model performs tool selection, structured argument generation, and answer synthesis over local deterministic tools with controlled fault injection.

Table~\ref{tab:model_in_loop_main} reports the main model-in-the-loop validation. All three methods complete the compact task set successfully. Therefore, this experiment is not primarily differentiating on final task success. Its value is as an external-validity bridge: it shows that the recovery-enabled orchestration machinery can run when a live model drives tool interaction, rather than only in a deterministic executor.

\begin{table}[t]
\centering
\caption{Main model-in-the-loop validation over 90 executions.}
\label{tab:model_in_loop_main}
\small
\begin{tabular}{lcccc}
\toprule
\textbf{Method} & \textbf{Executions} & \textbf{Success (\%)} & \textbf{Detection (\%)} & \textbf{Recovery Success (\%)} \\
\midrule
Retry-only & 30 & 100.0 & 46.7 & 20.0 \\
Full replanning & 30 & 100.0 & 46.7 & 46.7 \\
Self-healing & 30 & 100.0 & 46.7 & 46.7 \\
\bottomrule
\end{tabular}
\end{table}

The verifier ablation provides stronger live-model evidence for the role of verification. As shown in Table~\ref{tab:model_in_loop_verifier_ablation}, enabling verification improves task success from 96.7\% to 100.0\%, detection from 20.0\% to 46.7\%, and recovery success from 16.7\% to 46.7\% in the compact model-in-the-loop setting.

\begin{table}[t]
\centering
\caption{Model-in-the-loop verifier ablation over 60 executions.}
\label{tab:model_in_loop_verifier_ablation}
\small
\begin{tabular}{lcccc}
\toprule
\textbf{Variant} & \textbf{Executions} & \textbf{Success (\%)} & \textbf{Detection (\%)} & \textbf{Recovery Success (\%)} \\
\midrule
Verifier off & 30 & 96.7 & 20.0 & 16.7 \\
Verifier on & 30 & 100.0 & 46.7 & 46.7 \\
\bottomrule
\end{tabular}
\end{table}

The supplemental model-in-the-loop budget sweep reaches 100.0\% success for retry-only, full replanning, and self-healing at all tested budgets in this compact task set. Recovery success differs: retry-only remains at 25.0\%, while full replanning and self-healing reach 55.0\%. Because final success saturates, this supplemental sweep should not be treated as the primary evidence for budget efficiency. The controlled budget-sensitivity experiment in Section~\ref{subsec:budget_sensitivity_results} remains the main evidence that targeted recovery outperforms simply increasing recovery attempts. The model-in-the-loop sweep instead verifies that the budgeted recovery machinery executes correctly when a live model participates in tool use.

Overall, the model-in-the-loop results address an external-validity limitation of the controlled benchmark, but they do not constitute production validation. Tools remain local and deterministic, faults remain controlled, and the evaluation does not measure naturally occurring incidents, real external API reliability, or deployment-specific monetary cost.

\subsection{Trace-Based Diagnosability}
\label{subsec:trace_case_studies}

The controlled trace analysis extracts three representative self-healing traces: a runtime/tool failure, a context/evidence failure, and a semantic silent-failure case. Each trace records the task identifier, tool calls, injected fault, detected signal, inferred failure class, selected recovery action, verifier outcome, budget consumption, and final status. Table~\ref{tab:trace_case_studies} summarizes the three cases.

\begin{table}[t]
\centering
\caption{Representative self-healing recovery traces.}
\label{tab:trace_case_studies}
\footnotesize
\setlength{\tabcolsep}{5pt}
\begin{tabularx}{\textwidth}{L{0.22\textwidth} L{0.22\textwidth} L{0.24\textwidth} X}
\toprule
\textbf{Case} 
& \textbf{Task and Category} 
& \textbf{Fault} 
& \textbf{Recovery Outcome} \\
\midrule
Runtime/tool failure 
& \texttt{workflow\_001}; multi-step API workflow 
& Timeout 
& Retry with backoff; 1 verifier call; recovered: True \\
\midrule
Context/evidence failure 
& \texttt{retrieval\_001}; retrieval evidence 
& Stale context 
& Refresh retrieval; 1 verifier call; recovered: True \\
\midrule
Semantic silent failure 
& \texttt{retrieval\_002}; retrieval evidence 
& Semantic wrong output 
& Verification-guided recovery; 2 verifier calls; recovered: True \\
\bottomrule
\end{tabularx}
\end{table}

The traces support the observability claim: self-healing orchestration records not only final task status, but also why recovery was triggered and how the recovery decision was made. This makes recovery behavior inspectable for diagnosis, regression testing, and policy tuning.

\subsection{Summary of Findings}
\label{subsec:results_summary}

The results support seven findings. First, self-healing improves controlled runtime reliability, reaching 98.8\% success versus 94.5\% for retry-only and 93.8\% for full replanning. Second, the advantage grows under stress: self-healing maintains 97.3\% success in the high-stress regime, compared with 86.7\% for retry-only and 85.2\% for full replanning. Third, failure-type analysis shows that targeted recovery is strongest across all seven evaluated controlled fault classes. Fourth, ablations show that root-cause classification and targeted recovery are important contributors to reliability. Fifth, the matched-budget sweep shows that self-healing is not merely ``trying more'': it outperforms retry-only and full replanning at every tested recovery budget. Sixth, verifier-guided recovery reduces silent failures under the controlled semantic fault model. Seventh, compact model-in-the-loop validation shows that the same orchestration mechanism can operate when a live tool-calling model participates in tool use, while production API validation remains future work.

\section{Discussion}
\label{sec:discussion}

The results support the central thesis that reliability in tool-augmented LLM systems is partly an orchestration problem. Model capability remains important, but task completion also depends on how the system handles tool failures, stale or contradictory context, invalid intermediate states, verifier rejections, and recovery termination. The proposed self-healing orchestrator improves reliability in the controlled benchmark by treating these events as control-plane signals rather than as undifferentiated execution failures.

\subsection{Targeted Recovery Under Bounded Budgets}
\label{subsec:targeted_recovery_discussion}

The results show that different failure classes benefit from different recovery actions. Retry is useful for transient failures, but it is poorly matched to stale retrieval context, schema drift, contradictory evidence, or wrong-but-plausible semantic outputs. Full replanning is more general, but it can discard useful partial state and introduce unnecessary overhead when local repair, retrieval refresh, tool substitution, or verification would be sufficient. This explains why self-healing performs best in the failure-type analysis: the policy maps the inferred failure class to a recovery action that is more closely aligned with the likely cause.

The budget-sensitivity experiment is especially important for interpreting the result. The relaxed-budget ablation shows that additional recovery opportunities can improve completion, but the matched-budget sweep is more decisive: self-healing outperforms retry-only and full replanning at every tested recovery budget. The largest gap appears when the budget is most constrained. With one recovery attempt, self-healing reaches 94.0\% success, compared with 85.3\% for retry-only and 88.2\% for full replanning. This indicates that the improvement is not merely due to making more attempts; targeted recovery uses limited recovery opportunities more effectively.

At the same time, retry-only remains a useful baseline. For simple transient faults and low-risk tasks, retry may be the best first recovery action because it is cheap and easy to operate. The implication is not that retry should be removed, but that it should be one action within a broader recovery policy. A reliability-oriented orchestrator should decide when retry is appropriate, when local repair is sufficient, when replanning is necessary, and when degradation or escalation is safer than continued automatic recovery.

\subsection{Verification, Silent Failures, and Observability}
\label{subsec:verification_observability_discussion}

The semantic silent-failure experiment highlights a limitation of runtime monitoring alone. Many failures expose explicit signals, such as timeouts, malformed responses, or schema errors. Semantic failures are harder: the tool output or final answer can be syntactically valid and still be unsupported, stale, contradictory, or incorrect. In the controlled semantic fault setting, non-verifying methods return silent failures between 13.2\% and 17.6\% of the time, while verifier-guided self-healing reduces silent failure to 0.0\% under the evaluated templates.

This result supports treating verification as part of the control loop rather than as a final formatting check. A verifier rejection can be converted into a recoverable signal, allowing the orchestrator to refresh evidence, regenerate an answer under evidence constraints, or escalate when confidence remains too low. However, this result should not be read as a universal guarantee. Verifier quality depends on task definition, evidence availability, domain complexity, model behavior, and verifier design. In deployment, verification policies must be tuned to the risk of the task and the cost of false acceptance or false rejection.

Observability is complementary to verification. The no-observability ablation preserves task success but removes trace events, showing that traces primarily support diagnosability rather than immediate task completion in the controlled benchmark. This distinction matters operationally. Trace records make it possible to inspect failure signals, inferred classes, recovery actions, verifier outcomes, budget consumption, and final status. Such traces can support debugging, regression testing, incident review, policy tuning, and conversion of observed failures into future benchmark cases.

\subsection{Controlled Evaluation and Model-in-the-Loop Scope}
\label{subsec:model_in_loop_scope_discussion}

The controlled benchmark is designed for internal validity. Deterministic tools, fixed task definitions, matched fault schedules, and paired seeds make it possible to attribute differences in performance to orchestration behavior. This is useful for isolating the effect of failure classification, targeted recovery, verification, budgets, and observability. The limitation is that controlled experiments do not capture all sources of variability in real model and API behavior.

The model-in-the-loop validation partially addresses this gap. In that setting, a live tool-calling model performs tool selection, argument generation, and answer synthesis while tools remain local and fault-injected. The main model-in-the-loop experiment saturates on final success, so it should not be interpreted as the strongest differentiating result. Its value is that the same recovery mechanism operates when a live model participates in execution. The verifier ablation is more informative: enabling verification improves success, detection, and recovery behavior in the compact model-in-the-loop setting.

This evidence changes the scope of the claim. Live-model validation is no longer purely future work, but the current validation remains compact. It does not measure production API reliability, naturally occurring incidents, real user traffic, large-scale retrieval systems, or provider-specific monetary cost. The controlled benchmark should therefore be interpreted as the primary mechanism-isolation study, while the model-in-the-loop experiments provide a bridge toward live-model behavior.

\subsection{Deployment Implications}
\label{subsec:deployment_implications}

Deploying self-healing orchestration requires tuning recovery policies to application risk, latency constraints, tool cost, and safety requirements. The controlled cost proxy shows relative orchestration overhead, but it does not estimate real monetary cost. In practice, verifier calls may be expensive if they use a stronger model, long evidence context, or multiple cross-checking steps. Recovery policies should therefore control verifier frequency, recovery depth, model escalation, and human escalation thresholds.

The appropriate policy also depends on tool semantics. Read-only tools can often tolerate retry, retrieval refresh, or substitution. Tools with irreversible side effects, such as purchases, account changes, external notifications, or write operations, require more conservative recovery, stronger verification, idempotency checks, and possibly human confirmation. Authentication failures, authorization boundaries, privacy constraints, and audit requirements may also restrict which recovery actions are allowed.

The broader design implication is that reliable tool-augmented agents should combine failure-aware recovery, explicit budgets, verification, and traceability. The goal is not to maximize automatic recovery at any cost, but to recover when the system has enough evidence to proceed safely and to degrade, clarify, or escalate when it does not.

\section{Limitations and Threats to Validity}
\label{sec:limitations}

This section clarifies the scope of the evidence. The controlled benchmark is designed to isolate orchestration behavior under reproducible task, tool, and fault conditions. The compact model-in-the-loop experiments add live model participation, but they still use local deterministic tools and controlled faults. The results therefore support claims about failure-aware orchestration under the evaluated conditions; they do not by themselves establish production reliability for arbitrary LLM agents, external APIs, user populations, or deployment environments.

\subsection{Construct Validity}
\label{subsec:construct_validity}

Construct validity concerns whether the study measures the concepts it claims to measure. The primary constructs are task success, recovery success, silent failure, cost overhead, latency overhead, and diagnosability. The controlled benchmark operationalizes these using task-specific success criteria, injected fault records, verifier outcomes, simulated latency, call-count cost proxies, and trace events.

The benchmark improves construct validity by defining expected answers and success criteria for each task. However, synthetic success criteria are simpler than many real user goals. A response can satisfy a benchmark criterion while still lacking nuance, contextual appropriateness, or domain-specific safety requirements. Similarly, recovery success is measured relative to injected faults and expected outputs, not against full production user satisfaction or business correctness.

Cost and latency are also approximations. The controlled experiments use relative call-count proxies rather than monetary cost or wall-clock service latency. This enables consistent comparison across orchestration strategies, but it does not capture provider-specific pricing, token lengths, model tier differences, network variability, storage cost, or human-review cost. In deployment, verifier calls may dominate cost or latency if they use a stronger model, long evidence context, or multiple cross-checking steps.

Silent failure is measured using controlled wrong-but-plausible semantic faults. This isolates an important class of errors, but it does not cover all open-world silent failures. Real systems may involve ambiguous user intent, incomplete ground truth, adversarial prompts, missing evidence, multi-hop reasoning errors, or verifier mistakes. Results that reach 100.0\% within the evaluated templates should therefore be read as controlled benchmark outcomes, not as guarantees of universal recovery or silent-failure elimination.

\subsection{Internal Validity}
\label{subsec:internal_validity}

Internal validity concerns whether observed differences are caused by orchestration strategy rather than confounding factors. The controlled experiments mitigate this threat by using the same tasks, deterministic tools, fault-injection protocol, paired seed schedules, timeout thresholds, execution limits, and success criteria across methods. This paired design makes it more likely that observed differences reflect orchestration behavior rather than differences in task distribution or fault exposure.

Two internal-validity threats remain. First, the benchmark tasks and failure templates are author-constructed, which may unintentionally align them with the proposed recovery policy. We mitigate this by including multiple task categories, multiple detectable fault classes, semantic wrong-output faults, high-stress settings, ablations, and matched-budget experiments. Independent benchmark construction and external replication would strengthen this evidence further.

Second, the baselines are controlled implementations of broad strategy families rather than exhaustive versions of every possible agent framework. A more sophisticated ReAct-style agent, planner, fallback system, or verifier-equipped baseline could perform differently. The purpose of the comparison is therefore not to claim dominance over every possible implementation, but to evaluate whether explicit failure classification, targeted recovery, budgets, verification, and traceability improve reliability over representative static, retry-only, ReAct-style, and full-replanning strategies under matched conditions.

\subsection{External Validity}
\label{subsec:external_validity}

External validity concerns whether the results generalize beyond the evaluated setting. The primary controlled benchmark uses synthetic tasks, simulated tools, deterministic execution, and controlled fault injection. This design is appropriate for mechanism isolation, but it does not capture the full variability of production LLM applications, external APIs, retrieval systems, user behavior, authorization systems, network conditions, or side-effecting tool execution.

The model-in-the-loop experiments partially address this limitation by allowing a live tool-calling model to perform tool selection, argument generation, and answer synthesis. However, the validation remains compact. The tools are local and deterministic, faults are controlled, and the task set is much smaller than the controlled benchmark. These experiments provide a bridge to live model behavior, but they do not measure naturally occurring production incidents, real external API reliability, large-scale retrieval behavior, or real user traffic.

Recovery effectiveness may also vary by domain. A timeout may be safe to retry, but an authentication error may require re-authorization. Retrieval refresh may help stale evidence, but not ambiguous user intent. Tool substitution may be valid only when substitute tools have compatible semantics, authorization boundaries, and side-effect behavior. High-risk domains such as finance, healthcare, legal workflows, or operational automation may require stricter verification, conservative degradation, human escalation, or explicit user confirmation.

\subsection{Conclusion Validity and Reproducibility}
\label{subsec:conclusion_validity_reproducibility}

Conclusion validity concerns whether the conclusions follow from the collected data. The controlled main experiment contains 9000 executions; the controlled ablation contains 9000 executions; the controlled budget-sensitivity experiment contains 7200 executions; and the semantic silent-failure experiment contains 7200 executions. The model-in-the-loop experiments add 90 main-validation executions, 60 verifier-ablation executions, and 180 budget-sensitivity executions. These execution counts support stable comparison within the constructed benchmark, but they should not be interpreted as statistical estimates over the full population of real-world agent workloads.

The primary conclusion is comparative and scoped: under matched controlled tasks, fault schedules, and success criteria, self-healing orchestration improves reliability over the evaluated baselines and uses matched recovery budgets more effectively. The model-in-the-loop results support a narrower claim: the same orchestration mechanism can operate when a live model participates in tool use over local fault-injected tools. Broader claims about production reliability require additional evaluation.

The artifact package supports reproducibility by including controlled and model-in-the-loop notebooks, structured task definitions, result CSV files, generated figures, LaTeX tables, health checks, and JSON recovery traces. Reproducing the controlled numbers requires the same task definitions, deterministic executor, fault-injection code, and seed schedule. Extending the evaluation to external APIs or live production systems would require additional controls for model versioning, prompt changes, service availability, API changes, authorization state, and external data drift.

\subsection{Summary of Limitations}
\label{subsec:limitations_summary}

Table~\ref{tab:validity_threats_summary} summarizes the main limitations and mitigations.

\begin{table}[t]
\centering
\caption{Summary of validity threats and mitigation strategies.}
\label{tab:validity_threats_summary}
\small
\begin{tabularx}{\textwidth}{L{0.22\textwidth} L{0.38\textwidth} Y}
\toprule
\textbf{Threat} & \textbf{Limitation} & \textbf{Mitigation} \\
\midrule
Benchmark scope & The controlled benchmark uses 100 synthetic tasks. & Tasks span five tool-augmented categories and are released as structured artifacts. \\
\midrule
Task-generation bias & Author-constructed tasks may align with the proposed recovery policy. & We evaluate multiple fault classes, baselines, ablations, semantic faults, stress settings, and matched budgets. \\
\midrule
Simulated tools & Controlled tools do not capture all real API behavior, latency, authorization, or side effects. & Simulation enables reproducible fault injection and paired comparison across methods. \\
\midrule
Model-in-loop scale & Live-model validation is compact and uses local deterministic tools. & It provides a bridge to live model behavior while preserving controlled fault analysis. \\
\midrule
Fault-model completeness & The evaluated faults do not cover all production failures. & The benchmark includes runtime, output, context, control-loop, and semantic fault classes. \\
\midrule
Cost and latency & Reported overheads are relative proxies, not production dollar cost or wall-clock latency. & The same accounting scheme is applied consistently across methods; production cost remains future work. \\
\midrule
Verifier scope & The semantic verifier uses controlled task success criteria. & Results are framed as controlled silent-failure evidence, not a universal guarantee. \\
\midrule
Baseline implementation & Baselines are representative controlled implementations, not all possible optimized agents. & All methods share tasks, tools, seeds, faults, and success criteria for paired comparison. \\
\midrule
External generalization & Results may not directly transfer to production APIs, user traffic, or high-risk domains. & Broader validation requires real tools, production traces, larger task sets, and domain-specific policies. \\
\bottomrule
\end{tabularx}
\end{table}

Overall, these limitations define the scope of the contribution. The experiments provide controlled evidence that failure-aware, budgeted, verifier-guided orchestration improves reliability and diagnosability under the evaluated fault models. The compact model-in-the-loop validation strengthens external validity, but full production validation remains future work.

\section{Future Work}
\label{sec:future_work}

The most important next step is to extend the compact model-in-the-loop validation to larger live-model evaluations with real external APIs, production retrieval systems, realistic tool latency distributions, and naturally occurring failure traces. The controlled benchmark in this paper isolates orchestration behavior, and the model-in-the-loop experiments provide an initial bridge to live model behavior. Future evaluations should test whether the same reliability gains persist under changing model versions, stochastic tool-call generation, authentication and authorization failures, API schema drift, data freshness issues, and long-running workflows.

A second direction is to improve recovery policy learning and verification. The policy evaluated in this paper is intentionally interpretable and fixed so that recovery behavior is reproducible. Future work could study hybrid policies that preserve rule-based safety constraints while learning when to retry, repair arguments, refresh retrieval, substitute tools, replan, escalate, or terminate based on historical traces and recovery outcomes. Such policies should be evaluated not only for task success, but also for latency, token usage, provider-specific monetary cost, verifier overhead, and escalation burden. Stronger verification is also needed, including verifier calibration, evidence grounding, false-acceptance and false-rejection analysis, cross-source consistency checking, and human review for high-risk cases.

A third direction is to make self-healing safe for side-effecting and high-risk tool use. Recovery policies for read-only retrieval tasks can be more permissive than policies for workflows that modify accounts, send notifications, make purchases, update records, or trigger operational actions. Future systems should incorporate idempotency checks, authorization boundaries, privacy constraints, safety budgets, explicit user confirmation, and human escalation when automatic recovery cannot establish sufficient confidence. Finally, standardized observability schemas for agentic traces would make it easier to compare orchestrators, reproduce failures, build regression suites, and construct shared datasets of agent failures and recovery outcomes.

\section{Conclusion}
\label{sec:conclusion}

This paper presented a self-healing orchestrator for tool-augmented LLM agents, framing reliability as a runtime orchestration-control problem rather than only as a model-level capability. The proposed design maps execution state, failure signals, inferred failure classes, recovery budgets, verification outcomes, and observability traces into a structured control loop for bounded recovery.

Across the controlled fault-injection benchmark, self-healing achieved the highest overall reliability, reaching 98.8\% task success compared with 94.5\% for retry-only and 93.8\% for full replanning. The advantage was larger in the high-stress regime, where self-healing maintained 97.3\% success compared with 86.7\% for retry-only and 85.2\% for full replanning. Failure-type analysis and ablations showed that root-cause classification and targeted recovery are important contributors to this improvement. A matched recovery-budget sweep further showed that the gains are not merely due to making more attempts: self-healing outperformed retry-only and full replanning at every tested budget, with the largest gap when only one recovery attempt was available.

The semantic silent-failure experiment showed that verification is necessary for failures that are syntactically valid but semantically wrong. Under the controlled semantic fault model, verifier-guided self-healing reduced silent failures to 0.0\%, while non-verifying methods returned wrong-but-plausible outputs more often. This result should be interpreted within the evaluated semantic templates, not as a universal guarantee against silent failure. Finally, the compact model-in-the-loop validation showed that the same orchestration mechanism can operate when a live tool-calling model performs tool selection, argument generation, and answer synthesis over local fault-injected tools.

Overall, these results support self-healing orchestration as a practical control-plane pattern for improving reliability and diagnosability in tool-augmented LLM systems. The evidence is strongest within the controlled benchmark and compact model-in-the-loop setting; larger live-model studies with real APIs, production traces, naturally occurring failures, and measured deployment costs remain important future work.

\FloatBarrier

\section*{Acknowledgments}

The author thanks colleagues and reviewers of earlier drafts for helpful feedback. This work was conducted in the author's personal capacity. The views and conclusions expressed in this paper are solely those of the author and do not necessarily reflect the views of the author's employer.

\section*{Funding}

This work received no external funding.

\section*{Conflicts of Interest}

The author declares no conflicts of interest.

\section*{Artifact Availability}

The benchmark tasks, controlled and model-in-the-loop notebooks, result files, generated figures, LaTeX tables, health checks, and representative recovery traces are available in the public artifact repository:
\href{https://github.com/R-Suresh/self-healing-agentic-orchestrator}{GitHub repository}.

\bibliographystyle{plain}
\bibliography{references}

@inproceedings{lewis2020retrieval,
  author    = {P. Lewis and E. Perez and A. Piktus and F. Petroni and V. Karpukhin and N. Goyal and H. K{\"u}ttler and M. Lewis and W. Yih and T. Rockt{\"a}schel and S. Riedel and D. Kiela},
  title     = {Retrieval-Augmented Generation for Knowledge-Intensive {{NLP}} Tasks},
  booktitle = {Adv. Neural Inf. Process. Syst.},
  volume    = {33},
  pages     = {9459--9474},
  year      = {2020},
  doi       = {10.48550/arXiv.2005.11401}
}

@inproceedings{yao2023react,
  author    = {S. Yao and J. Zhao and D. Yu and N. Du and I. Shafran and K. Narasimhan and Y. Cao},
  title     = {{{ReAct}}: Synergizing Reasoning and Acting in Language Models},
  booktitle = {Proc. Int. Conf. Learn. Represent. (ICLR)},
  year      = {2023},
  url       = {https://openreview.net/forum?id=WE_vluYmi-W}
}

@inproceedings{schick2023toolformer,
  author    = {T. Schick and J. Dwivedi-Yu and R. Dess{\`i} and R. Raileanu and M. Lomeli and L. Zettlemoyer and N. Cancedda and T. Scialom},
  title     = {{{Toolformer}}: Language Models Can Teach Themselves to Use Tools},
  booktitle = {Adv. Neural Inf. Process. Syst.},
  volume    = {36},
  pages     = {68539--68551},
  year      = {2023}
}

@inproceedings{qin2024toolllm,
  author    = {Y. Qin and S. Liang and Y. Ye and K. Zhu and L. Yan and Y. Lu and Y. Lin and X. Cong and X. Tang and B. Qian and S. Zhao and L. Hong and R. Tian and R. Xie and J. Zhou and M. Gerstein and D. Li and Z. Liu and M. Sun},
  title     = {{{ToolLLM}}: Facilitating Large Language Models to Master 16000+ Real-World {{APIs}}},
  booktitle = {Proc. Int. Conf. Learn. Represent. (ICLR)},
  year      = {2024},
  url       = {https://openreview.net/forum?id=S63InEQU9N}
}

@article{wang2024surveyagents,
  author  = {L. Wang and C. Ma and X. Feng and Z. Zhang and H. Yang and J. Zhang and Z. Chen and J. Tang and X. Chen and Y. Lin and W. X. Zhao and Z. Wei and J. R. Wen},
  title   = {A Survey on Large Language Model Based Autonomous Agents},
  journal = {Front. Comput. Sci.},
  volume  = {18},
  number  = {6},
  pages   = {186345},
  year    = {2024},
  doi     = {10.1007/s11704-024-40231-1}
}

@inproceedings{shinn2023reflexion,
  author    = {N. Shinn and F. Cassano and E. Berman and A. Gopinath and K. Narasimhan and S. Yao},
  title     = {Reflexion: Language Agents with Verbal Reinforcement Learning},
  booktitle = {Adv. Neural Inf. Process. Syst.},
  volume    = {36},
  pages     = {8630--8652},
  year      = {2023}
}

@inproceedings{liu2024agentbench,
  author    = {X. Liu and H. Yu and H. Zhang and Y. Xu and X. Lei and H. Lai and Y. Gu and H. Ding and K. Men and K. Yang and S. Zhang and X. Deng and A. Zeng and Z. Du and C. Zhang and S. Shen and T. Zhang and Y. Su and H. Sun and M. Huang and Y. Dong and J. Tang},
  title     = {{{AgentBench}}: Evaluating {{LLMs}} as Agents},
  booktitle = {Proc. Int. Conf. Learn. Represent. (ICLR)},
  year      = {2024},
  url       = {https://openreview.net/forum?id=z8HeY9rq9S}
}

@inproceedings{winston2025taxonomy,
  author    = {C. Winston and R. Just},
  title     = {A Taxonomy of Failures in Tool-Augmented {{LLMs}}},
  booktitle = {Proc. IEEE/ACM Int. Conf. Autom. Softw. Test (AST)},
  pages     = {125--135},
  year      = {2025},
  doi       = {10.1109/AST66626.2025.00019}
}

@article{gupta2026reliabilitybench,
  author  = {A. Gupta},
  title   = {{{ReliabilityBench}}: Evaluating {{LLM}} Agent Reliability Under Production-Like Stress Conditions},
  journal = {arXiv preprint arXiv:2601.06112},
  year    = {2026}
}

@article{kephart2003vision,
  author    = {Kephart, Jeffrey O. and Chess, David M.},
  title     = {The Vision of Autonomic Computing},
  journal   = {Computer},
  volume    = {36},
  number    = {1},
  pages     = {41--50},
  year      = {2003},
  doi       = {10.1109/MC.2003.1160055},
  publisher = {IEEE}
}

@techreport{ibm2006autonomic,
  author      = {{IBM Corporation}},
  title       = {An Architectural Blueprint for Autonomic Computing},
  institution = {IBM Corporation},
  year        = {2006},
  type        = {White paper}
}

\end{document}